\documentclass[lettersize,journal]{IEEEtran}
\usepackage{amsmath,amsfonts}
\usepackage{algorithmic}
\usepackage{algorithm}
\usepackage{array}
\usepackage[caption=false,font=normalsize,labelfont=sf,textfont=sf]{subfig}
\usepackage{textcomp}
\usepackage{stfloats}
\usepackage{amsmath,graphicx,multirow,booktabs}
\usepackage{url}
\usepackage{verbatim}
\usepackage{graphicx}
\usepackage{cite}
\usepackage{xcolor}
\usepackage{colortbl}
\usepackage{arydshln}
\usepackage{enumerate}

\begin{document}

\title{A Deep Semantic Segmentation Network with Semantic and Contextual Refinements}

\author{Zhiyan Wang, Deyin Liu, Lin Yuanbo Wu,~\IEEEmembership{Senior Member,~IEEE,} Song Wang*, Xin Guo, Lin Qi
\thanks{Zhiyan Wang, Song Wang, Xin Guo, and Lin Qi are with the School of Electrical and Information Engineering, Zhengzhou University, Zhengzhou 450001, China (e-mail: wangsong61@163.com). Deyin Liu is with the School of Artificial Intelligence, Anhui University, Hefei 230039, China. Lin Yuanbo Wu is with Department of Computer Science, Swansea University, SA1 8EN, UK.}
}

\markboth{Journal of \LaTeX\ Class Files,~Vol.~14, No.~8, August~2021}%
{Shell \MakeLowercase{\textit{et al.}}: A Sample Article Using IEEEtran.cls for IEEE Journals}

\IEEEpubid{0000--0000/00\$00.00~\copyright~2021 IEEE}

\maketitle

\begin{abstract}
Semantic segmentation is a fundamental task in multimedia processing, which can be used for analyzing, understanding, editing contents of images and videos, among others. To accelerate the analysis of multimedia data, existing segmentation researches tend to extract semantic information by progressively reducing the spatial resolutions of feature maps. However, this approach introduces a misalignment problem when restoring the resolution of high-level feature maps. In this paper, we design a Semantic Refinement Module (SRM) to address this issue within the segmentation network. Specifically, SRM is designed to learn a transformation offset for each pixel in the upsampled feature maps, guided by high-resolution feature maps and neighboring offsets. By applying these offsets to the upsampled feature maps, SRM enhances the semantic representation of the segmentation network, particularly for pixels around object boundaries. Furthermore, a Contextual Refinement Module (CRM) is presented to capture global context information across both spatial and channel dimensions. To balance dimensions between channel and space, we aggregate the semantic maps from all four stages of the backbone to enrich channel context information. The efficacy of these proposed modules is validated on three widely used datasets—Cityscapes, Bdd100K, and ADE20K—demonstrating superior performance compared to state-of-the-art methods. Additionally, this paper extends these modules to a lightweight segmentation network, achieving an mIoU of 82.5\% on the Cityscapes validation set with only 137.9 GFLOPs.
\end{abstract}


\section{Introduction}
\IEEEPARstart
{S}{emantic} segmentation, as a fundamental task in multimedia, involves assigning semantic labels to every pixel in an image. It has a variety of applications in multimedia data processing and analysis, including image/video editing \cite{ImageEditing1,ImageEditing2} and medical image analysis \cite{MedicalTMM, Medical1}, as well as in emerging industries such as autonomous driving \cite{Driving1,Drivingtmm}, etc. Extensive research \cite{DeepLab, Tao2020HierarchicalMA, CAEmm, CCNet, SRRNet} based on Full Convolution Network (FCN) \cite{FCN} has significantly advanced semantic segmentation since it was proposed by Long et al.. A typical FCN segmentation network follows an encoder-decoder architecture. The encoder extracts semantic representation from input images, while the decoder is responsible for label-level classification based on the encoder's output.

To enhance semantic representation, the encoder progressively reduces the spatial resolution of feature maps to expand the receptive fields of convolutional layers. While other techniques (such as large kernel convolutions \cite{peng2017large} and atrous convolutions \cite{DeepLab, zhu2019improving}) can also increase receptive fields, they often come with heavy computation burdens. The reduction of spatial resolution in feature maps significantly decreases the computational complexity of the encoder. However, this design introduces new challenges: 
(1) The missing semantic information from the downsampled feature maps causes a serious misalignment problem, negatively impacting the semantic label assignment for the pixels around object boundaries. (2) The receptive field is still confined to a limited region, making it hard to exploit the contributions of the out-of-region pixels to the semantic labeling. Accordingly, we present the related researches and their limitations with regard to each of the two challenges.

\IEEEpubidadjcol

\begin{figure}[tbp]
	\centering
    \includegraphics[width=\linewidth]{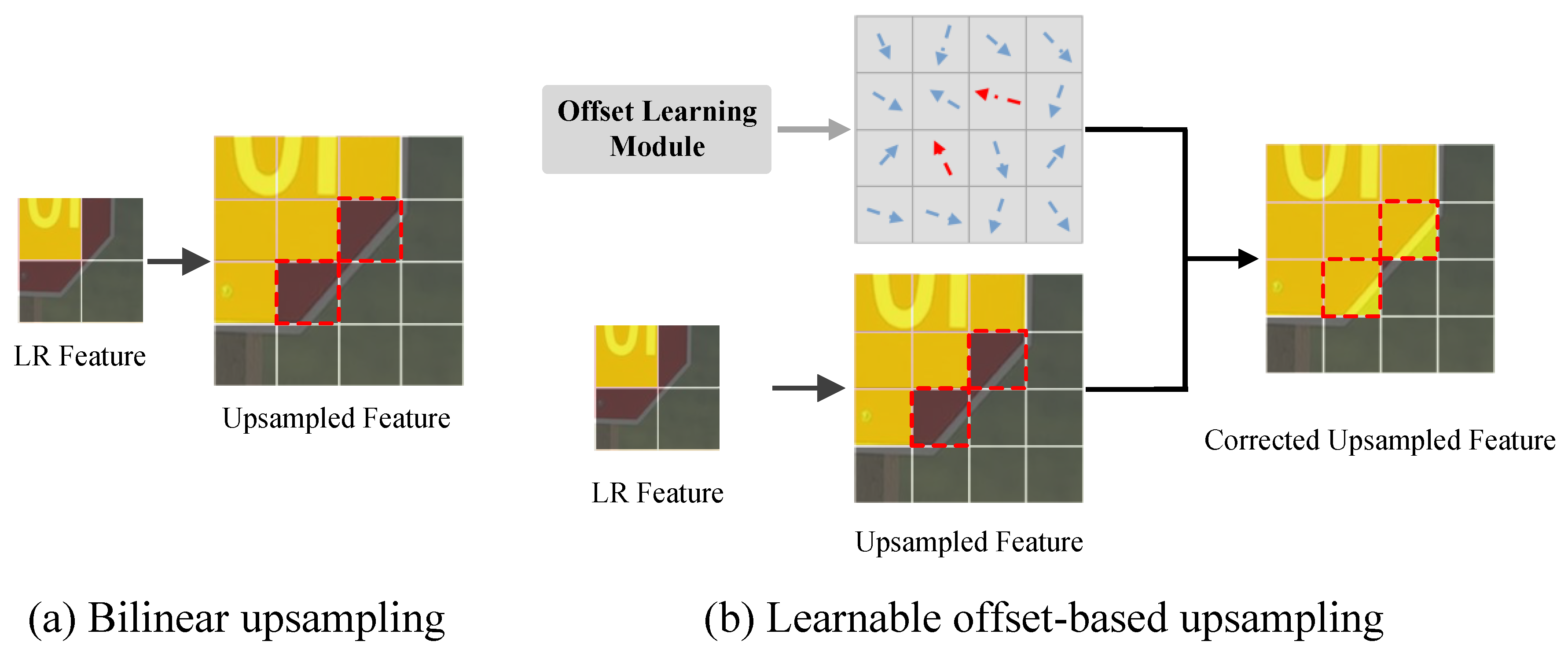}
    \vspace{-0.5cm}
	\caption{Comparison of Bilinear upsampling and Learnable offset-based upsampling for a Low-Resolution (LR) feature map.}
	\label{offsets}
 \vspace{-0.3cm}
\end{figure}

\begin{figure}[tbp]
	\centering
	\captionsetup[subfloat]{labelsep=none,format=plain,labelformat=empty,labelfont=rm,textfont=rm}
    \subfloat[{\scriptsize (a) w/o mask layer }] {\includegraphics[width=4.35cm]{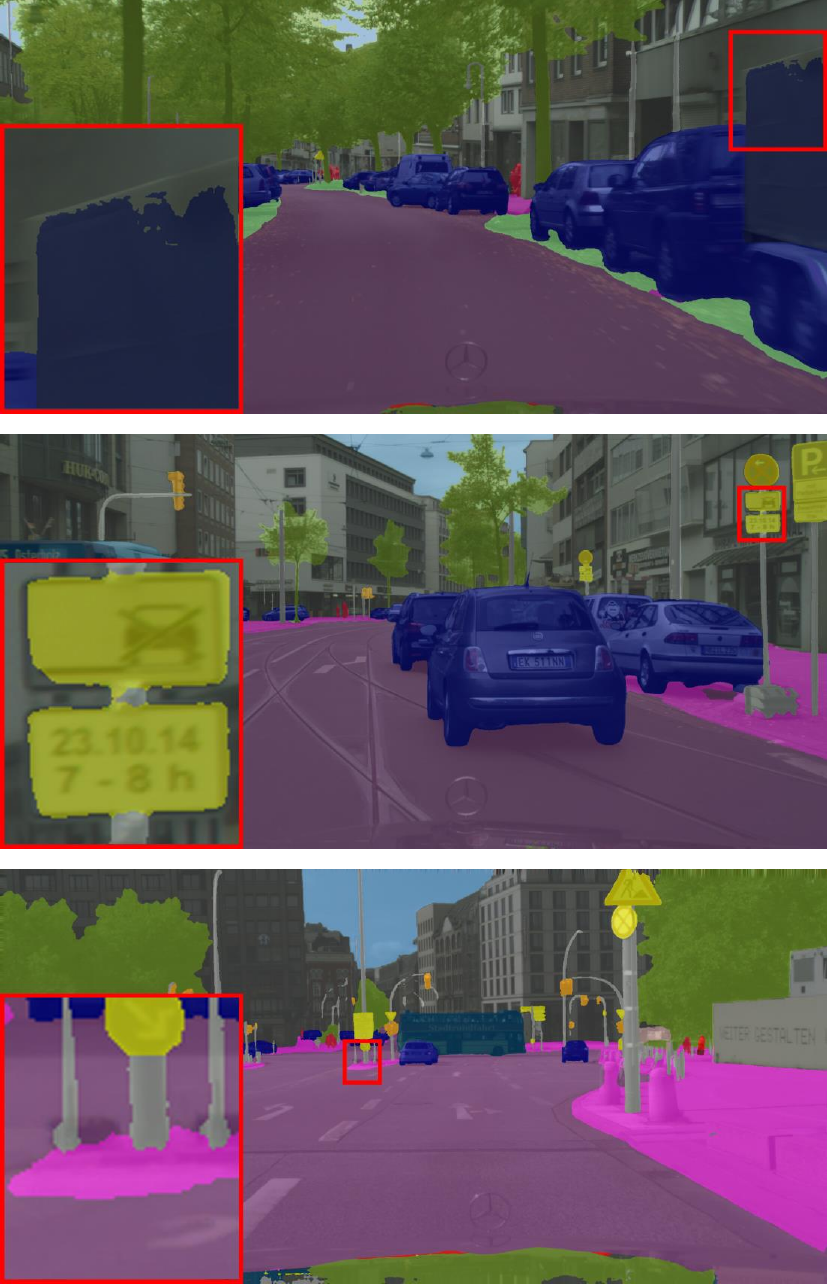}%
    }
    \hspace{0.02cm}
    \subfloat[{\scriptsize (b) w/ mask layer}]  {\includegraphics[width=4.35cm]{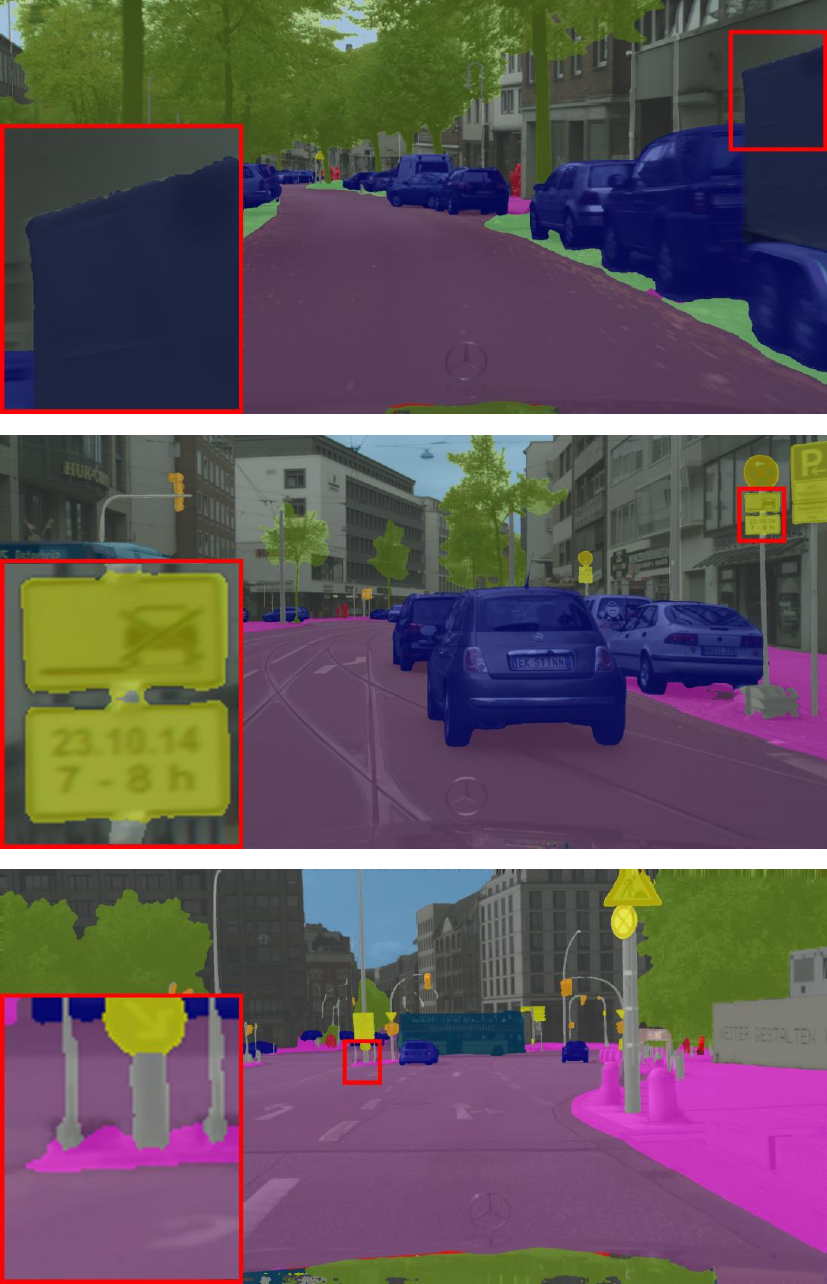}%
    }
  
	\caption{Some visualization comparisons on the Cityscapes dataset. (a) w/o mask layer, (b) w/ mask layer. The mask layer exploits the contribution of the neighbors' offsets to the ultimate offset for each pixel. The first column displays the outputs of the method without the ``mask" layer. The second column indicates the outputs of the method with the ``mask" layer. It is observed that the object boundaries generated by the ``mask" layer are clearer, such as the ``truck" in the first row and the ``traffic sign" in the second row. In the third row, the regions of ``sidewalk" and ``road" are classified more accurately when using the method with the ``mask" layer.}
	\label{mask_comparisons}
    \vspace{-0.5cm}
\end{figure}

Firstly, progressively downsampling the feature maps in the encoder indeed reduces the computation burden of the segmentation network. However, it is easy to cause the misalignment problem at object boundaries when the decoder recovers the resolution of semantic feature maps. The widely used upsampling methods, such as bilinear and nearest neighbor interpolation, struggle to obtain high-resolution feature maps with precise spatial semantic information \cite{GUM, Tian2019DecodersMF,LRD}.
Fig. \ref{offsets}(a) shows the bilinear upsampling process of a low-resolution feature map. The pixels around the boundaries of a traffic sign (with yellow mask) are assigned to the wrong class. In this case, recent research begins to learn an offset map for correcting the semantic classes of the pixels. Fig.\ref{offsets}(b) depicts a common offset-based upsampling process. The learned offset map adjusts the interpolating positions by adding bidimensional coordinates vectors to the regular ones in bilinear interpolation. In this case, the semantic class of the pixels around the boundaries of the traffic sign is corrected, as shown in the red boxes of Fig. \ref{offsets}(b). Specifically, GUM \cite{GUM} predicts a high-resolution guidance offset table to guide the upsampling process of low-resolution maps. AlignSeg \cite{AlignSeg} learns 2D transformation offsets for both high-resolution and low-resolution maps, aligning spatial features and reducing excessive details within objects, respectively. SFNet \cite{SFNet} interprets the difference between two feature maps of arbitrary resolutions from the same image as the motion of each pixel from one feature map to the other, introducing the concept of ``semantic flow". By learning the semantic flow between two network layers of different resolutions, the upsampled feature map aligns with spatial details in the high-resolution map. However, the aforementioned strategies learn transformation offsets by examining shifts of individual corresponding pixel pairs from feature maps with different resolutions, overlooking the influence of their neighbors in the local area. To address the above mentioned issue, this paper proposes a new solution by leveraging the contribution of neighbors' offsets to the ultimate offset learning for each pixel in upsampled feature maps. The visual results, as depicted in Fig. \ref{mask_comparisons}, illustrate the advantages of considering neighbors' offsets in improving the semantic label assignment for pixels around object boundaries.

\begin{figure}[tbp]
	\centering
	\captionsetup[subfloat]{labelsep=none,format=plain,labelformat=empty,labelfont=rm,textfont=rm}
    \subfloat[{\scriptsize (a) Average pooling }] {\includegraphics[width=2.95cm]{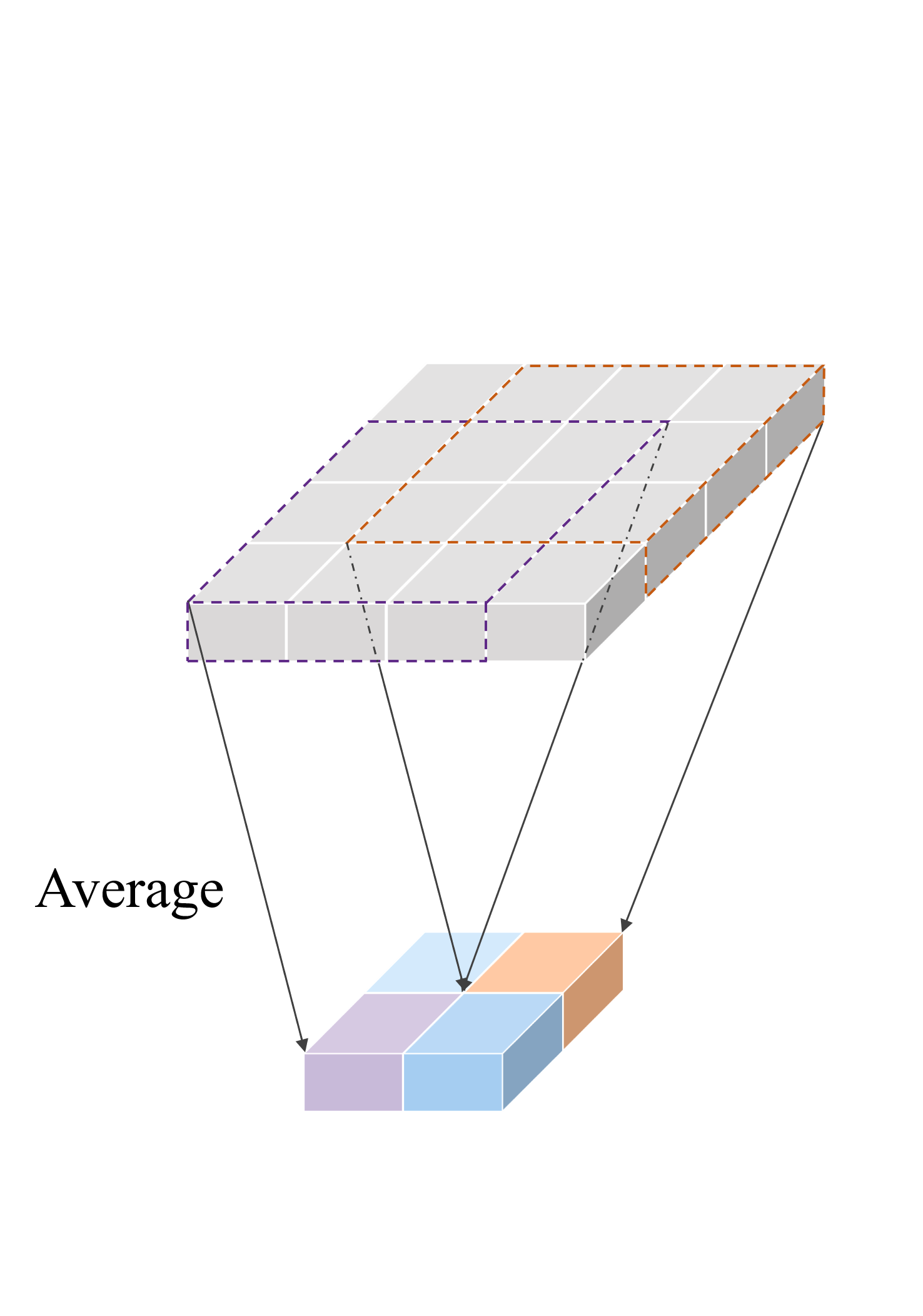}%
    }
    \subfloat[{\scriptsize (b) Attention mechanism }]  {\includegraphics[width=5.92cm]{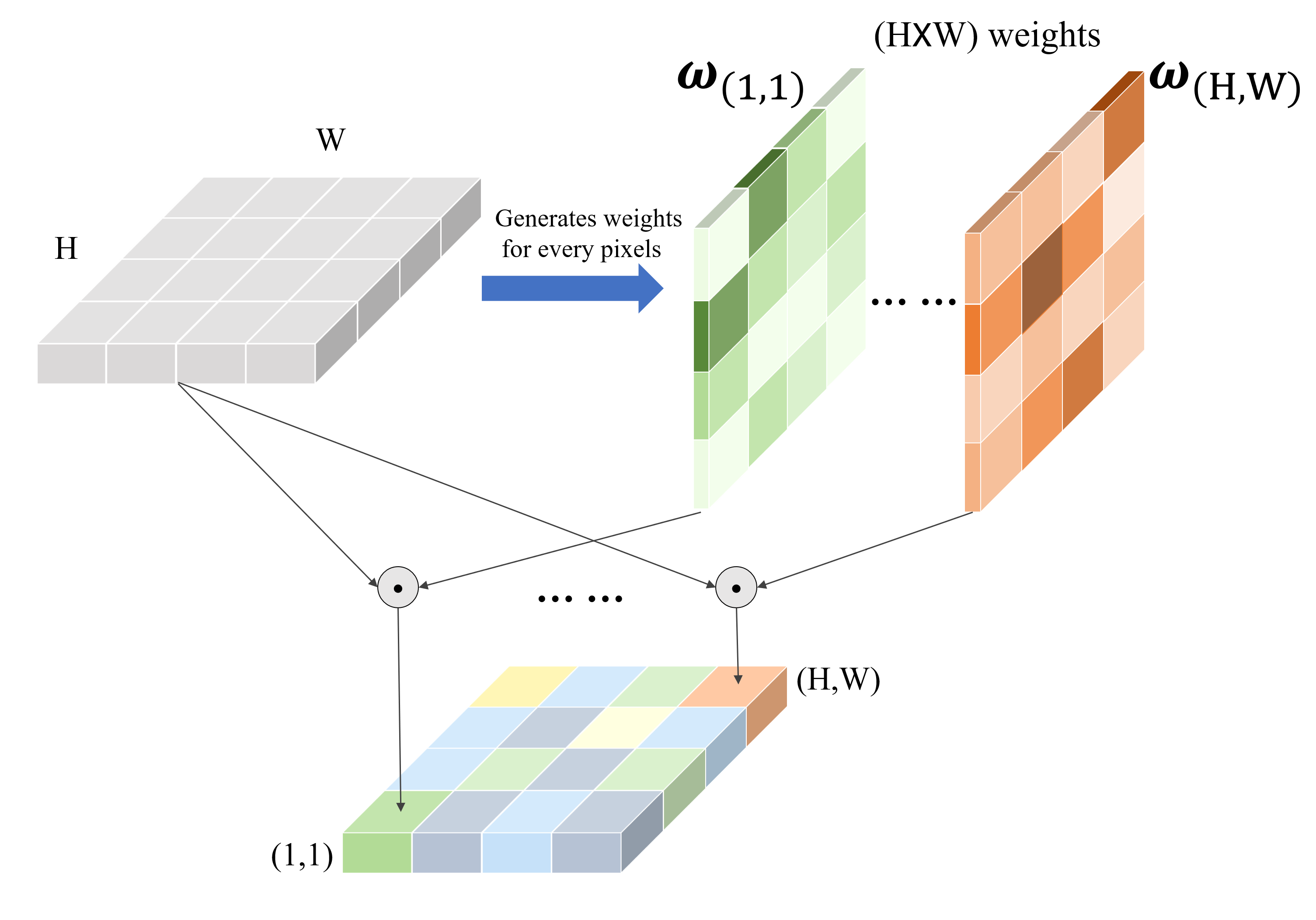}%
    }
	\caption{Comparison of the average pooling strategy and attention mechanism. (a) Average pooling strategy captures the contexts by taking the average of all pixels within the pooling region, overlooking the fact that different pixels may make unequal contributions. The pooling region for global average pooling encompasses the entire feature map. (b) Attention mechanism adaptively captures global contexts by calculating the response at a position through a weighted aggregation of features from all positions.}
	\label{context_comparisons}
 \vspace{-0.4cm}
\end{figure}

Secondly, the reduction of spatial resolution in feature maps also expands the receptive field of the segmentation network. However, its range is still limited. As the regions of objects from different categories in the image may have a similar appearance, local context information extracted from the limited receptive field is not enough to classify illegible pixels from different parts in the image. FCN-based methods typically employ a backbone network pre-trained on ImageNet to extract semantic features from the input images. The successive convolution and downsampling operations in the backbone expand the receptive field which reflects the context information of the input images. However, the backbone networks can only capture local context information from confined receptive fields and still fall short of including global context information. To enlarge the receptive field to explore context information, several studies perform pooling strategies on the semantic feature maps extracted from backbones. ParseNet \cite{ParseNet} introduces a global average pooling mechanism to derive global context by pooling features from the final stage of the backbone. As a result, ParseNet significantly improves the quality of semantic segmentation by amalgamating global features with local ones. Pyramid Pooling Module (PPM) \cite{PPM} is designed to exploit the capability of global context information through the fusion of context information at various pyramid scales. Deep Aggregation Pyramid Pooling Module (DAPPM) \cite{DDRNet} improves the approach by strengthening PPM \cite{PPM} with more scales and deep feature aggregation, providing richer context information. BSCNet \cite{Zhou2024BoundaryGuidedLS} designs a hierarchical pyramid pooling module to facilitate context fusion in a global-to-local manner. However, those pooling strategies only capture the contexts by taking the average of all pixels within the pooling region, overlooking the fact that different pixels in this area might contribute differently to the semantic labeling of a specific pixel, as illustrated in Fig. \ref{context_comparisons}(a). The attention mechanism offers a promising way to explore the contributions of different pixels to the semantic labeling (as shown in Fig. 3(b)). Zhao et al. \cite{zhao2018psanet} propose a point-wise spatial attention network to harvest contextual information from all positions in the feature map. Huang et al. \cite{CCNet} design a criss-cross attention module to capture contextual information from full-image spatial dependencies. SENet \cite{SENet} is designed to utilize the lightweight gating mechanism to model the context dependencies in the channel dimension. MCCA \cite{MCCA} proposes to perceive the contextual information from multiscale channels. However, most of attention-based algorithms focus on either the spatial relationship \cite{zhao2018psanet,CCNet} or the channel context \cite{SENet,MCCA}, ignoring that the semantic label assignment for each pixel is affected by the other pixels in both spatial and channel dimensions.

To this end, this paper introduces a Semantic and Contextual Refinement Network for segmentation tasks. The Semantic Refinement Module (SRM) is designed to precisely assign semantic labels to pixels around object boundaries in upsampled semantic feature maps. Initially, SRM utilizes high-resolution feature maps with rich spatial information from the encoder as guidance to learn a coarse transformation offset for every pixel in the upsampled semantic feature maps. Subsequently, SRM assesses the impact of neighbors' offsets in the local area on each pixel, employing a pixel-wise mask to refine the coarse transformation offset map. Additionally, to capture more meaningful global context information, the Contextual Refinement Module (CRM) incorporates an attention mechanism into the segmentation network. This mechanism sequentially explores dependencies between pixels across both spatial and channel dimensions. Acknowledging the imbalance between the spatial and channel dimensions, this paper aggregates semantic maps from all four stages of the backbone before extracting context information. In comparison to our previous conference paper \cite{wang2023feature}, this paper proposes a brand-new semantic refinement module to leverage the contribution of neighbors' offsets to offset map learning. Moreover, while the feature refinement module in the conference paper only explores the spatial dependencies between pixels, this paper proposes to capture global context information across both spatial and channel dimensions and builds a new contextual refinement module. To demonstrate the generality and effectiveness of the proposed modules, this paper further extends them to various segmentation networks, and analyzes their influence on the segmentation results.

In a summary, the contributions of this paper are given as follows:
\begin{enumerate}[(1)]
    \item We propose the Semantic Refinement Module which delves into the impact of neighbors' offsets in the local area on the shift of each pixel in upsampled semantic feature maps. It leads to a more precise semantic label assignment for pixels around object boundaries.
    \item The second innovative Contextual Refinement Module explores dependencies between pixels across both spatial and channel dimensions to capture global context information. To enhance channel context information, the paper aggregates semantic maps from all four stages of the backbone, thereby increasing the number of channels.
    \item To validate the effectiveness and generality of SRM and CRM, the proposed modules are applied to various segmentation networks, including lightweight ones. Extensive experiments demonstrate that these modules consistently achieve superior semantic segmentation performance, establishing new state-of-the-art results for Cityscapes, Bdd100K, and ADE20K.
\end{enumerate}

The paper is structured as follows. Section II provides a review of the related work. The architecture of the proposed network is described in Section III. In Section IV, we evaluate the performance of the proposed modules on three datasets. Conclusions are drawn in Section V.

\section{Related Work}
\subsection{Semantic Segmentation}
Semantic segmentation aims at assigning semantic labels for every pixel in the image. Based on FCN, a series of investigations have improved segmentation performance in many aspects. Recent works mainly focus on semantic feature enhancement \cite{U-Net,fpn,badrinarayanan2017segnet} and context modeling \cite{ParseNet,PPM,DDRNet,DeepLab}. 

\subsection{Semantic Feature Enhancement}
A direct and useful way to enhance the semantic features is by aggregating multi-level feature maps with different resolutions. For example, Unet \cite{U-Net} and FPN \cite{fpn} fuse the low-level feature with spatial details into the decoding stage via lateral connections, enhancing the representation power of the networks for semantic features. However, such an aggregation suffers from the semantic feature misalignment problem, where the upsampled low-resolution feature maps lack accurate spatial semantic information. In this case, current researches concentrate on designing learnable alignment strategies to boost the performance gain brought by the feature aggregation methods. SegNet \cite{badrinarayanan2017segnet} utilizes pooling indices determined during the max-pooling phase to recover more spatial details in the corresponding upsampling step. GUM \cite{GUM} employs the high-resolution feature map with more spatial details as a guidance to upsample the low-resolution one, reducing the spatial shift of the upsampled feature map. 
Unlike aligning feature maps at a fixed resolution, IFA \cite{hu2022learning} allows for the learning of features that accurately represent continuous fields of information, and achieves a precise feature alignment between adjacent feature maps. In HFGD \cite{Huang2023HighlevelFG}, high-level features from the backbone are used to guide the upsampler features towards more discriminative. SFNet \cite{SFNet} is proposed to learn a semantic flow map to represent the “motion” of every pixel between feature maps of adjacent levels. In AlignSeg \cite{AlignSeg}, the aligned feature aggregation module learns the transformation offsets for both low-resolution and high-resolution feature maps to reduce excessive details within objects. Instead of only relying on the different-resolution feature maps for guidance, this paper exploits the contribution of the neighbors' offsets to the ultimate offset for each pixel in the upsampled feature maps.

\subsection{Context Modeling}
Context information is able to provide the surrounding positional distribution relationship for pixels, which is crucial for semantic segmentation. To enlarge the receptive field of the context modeling, atrous convolution \cite{atr} is introduced into semantic segmentation to increase the size of the kernels with a limited computational cost. PAC \cite{PAC} suggests the concept of perspective-adaptive convolutions, which are utilized to capture receptive fields of diverse sizes and configurations. 
To break the limitation of the convolution operation on receptive fields, several studies implement spatial pooling at multiple scales to model context \cite{ParseNet,PPM,DDRNet,DeepLab, Zhou2024BoundaryGuidedLS }. However, these solutions only capture the contexts by taking the average of all pixels within the pooling region (as shown in Fig. \ref{context_comparisons}(a)), ignoring the difference of the contributions of the pixels in this pooling area to the context modeling. Thus, the attention mechanism is adopted by the current algorithms to model the global context information in an adaptive way (as shown in Fig. \ref{context_comparisons}(b)). PSANet \cite{zhao2018psanet} designs a pointwise spatial attention module that is designed to dynamically model long-range context information.
Wang et al. \cite{NL} propose a non-local block module to capture global dependencies by a weighted mean of all pixels across the spatial dimension. Yin et al. \cite{DNL} decouple the non-local block into a whitened term and a unary term, thereby facilitating attention computation. In addition, some works explore context information in the channel dimension. 
SENet \cite{SENet} employs a lightweight gating mechanism to model relationships between channels, which serves to enhance the network's representational capabilities. EncNet \cite{EncNet} builds a context encoding module that captures semantic encoding and predicts scaling factors, allowing for selective accentuation of class-dependent features. MCCA \cite{MCCA} proposes a multiscale channelwise cross attention network to perceive the context information of both large and small objects. There have been several researches focusing on the dual attention module to exploit context information across both spatial and channel dimensions. In \cite{HSNet}, a hierarchical feature refinement module is designed to refine features from both spatial and channel dimensions to alleviate the multilevel semantic gap. Nevertheless, the spatial attention in \cite{HSNet} is implemented by convolutional blocks, which are unable to directly capture global spatial context information. DANet \cite{DANet} designs a parallel framework to capture spatial and channel global dependencies respectively by employing the self-attention mechanism. However, its parallel design easily causes gradient information segregation, leading to a locally optimal solution. In this paper, CRM is proposed to explore the global context information across both spatial and channel dimensions in a series way. To achieve the global dependency assessment for each pixel on both dimensions, CRM builds a serial framework to connect channel attention and spatial attention. The serial framework is able to involve one attention in the gradient computation of the other attention during back-propagation, implementing the interactive learning between the attentions. Moreover, since the channel dimension is usually much less than spatial one, the proposed CRM aggregates semantic feature maps from each of the four backbone stages to enrich channel context information.

\begin{figure*} [tbp]	
    \centering
	\includegraphics[width= 0.95 \linewidth] {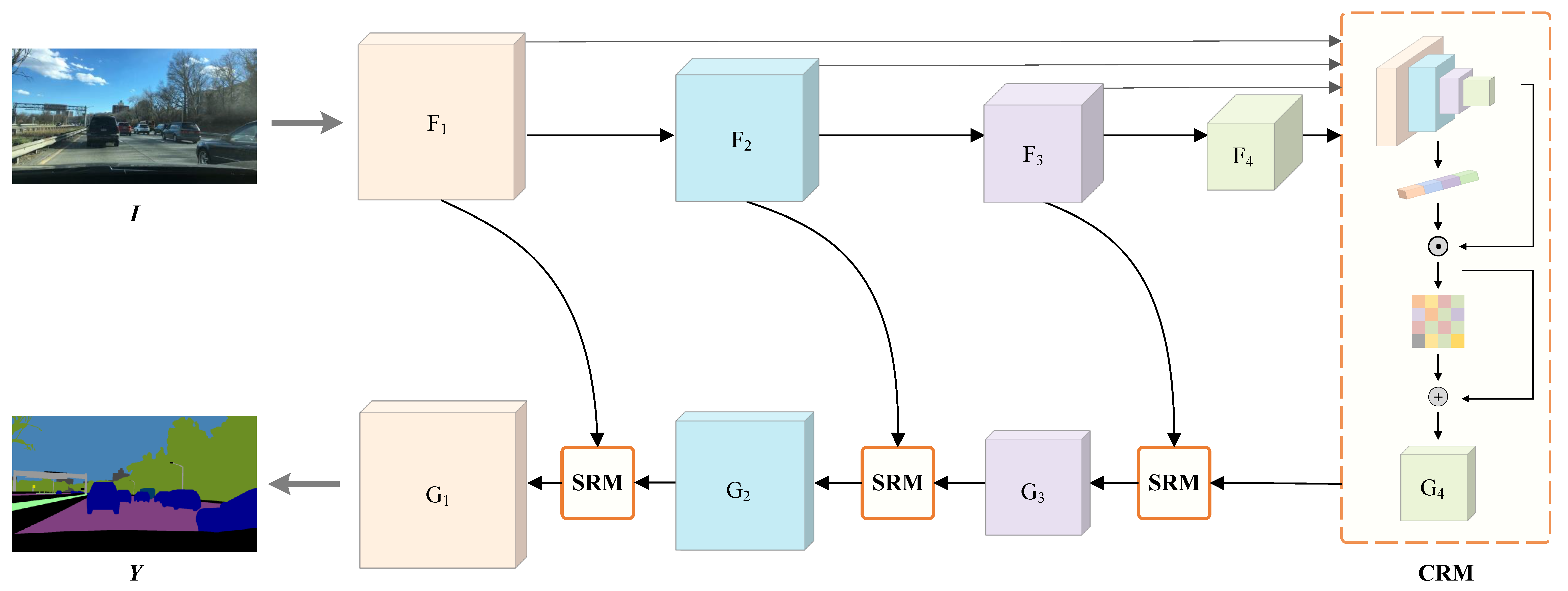}
	\caption{The structure of the proposed method. The Contextual Refinement Module takes all the four stages feature from the backbone as inputs to extract semantic features. The Semantic Refinement Module mitigates the feature misalignment problem during upsampling operations.}
 \label{whole_archi}
 \vspace{-0.3cm}
\end{figure*} 

\section{Method} 
In this section, we outline the details of the proposed segmentation network. First, we present the problem formulation and the overall network architecture of the proposed method in Sec. III-A and Sec. III-B. Then, Sec. III-C delves into the details of  SRM. CRM is presented in Sec. III-D. After that, Sec. III-E introduces the hybrid loss function employed in the proposed method.

\subsection{Problem Formulation}

Given an image $I \in\mathrm{\mathbb{R}} ^ {\ H  \times W \times 3  } $,  the semantic segmentation network $ \mathcal{F} $ aims at predicting a segmentation map $Y \in\mathrm{\mathbb{R}} ^ {\ H \times W}$, where each pixel of the image is assigned a semantic label. $H$ and $W$ indicate the height and width of the input image separately. This prediction process is represented in Eq. (1).
\begin{equation}
 Y = \mathcal{F}(I;\theta),
\end{equation}
where $\theta$ denotes the learnable parameters in the segmentation network $\mathcal{F}$. Specifically, the semantic segmentation network $ \mathcal{F}$ can be further divided into encoder, CRM, and decoder with SRM, as clarified in Sec. III-B. To obtain the desired prediction result, the optimization objective of network parameters $\theta$ is constructed to minimize the distance between the predicted segmentation map $Y$ and the corresponding ground truth $Y_G$
\begin{equation}
\mathop{\min}_{\theta} || Y - Y_G  ||.
\end{equation}

\subsection{Network Architecture} 
The proposed semantic segmentation network is founded on an encoder-decoder structure, including encoder, CRM, and decoder, as illustrated in Fig. \ref{whole_archi}. The encoder comprises a backbone network, aiming at extracting the semantic features from the input images. There are four stages in the backbone, which are denoted as $\left\{F_l\right\}_{l=1,...,4}$. The output size of $F_l$ is $\ \frac{H}{2^{l+1}} \times \frac{W}{2^{l+1}}\ $, where $H$ and $W$ are the input image height and width respectively. 
 CRM is designed to extract the global context information from the multi-stage features generated by the backbone. The role of the decoder is to translate the extracted semantics into the labels for each pixel, where SRM is designed to replace normal bilinear upsampling in the decoder of feature pyramid network \cite{fpn} to improve label prediction accuracy.

 \begin{figure}[tbp]
	\centering
	\includegraphics[width=\linewidth]{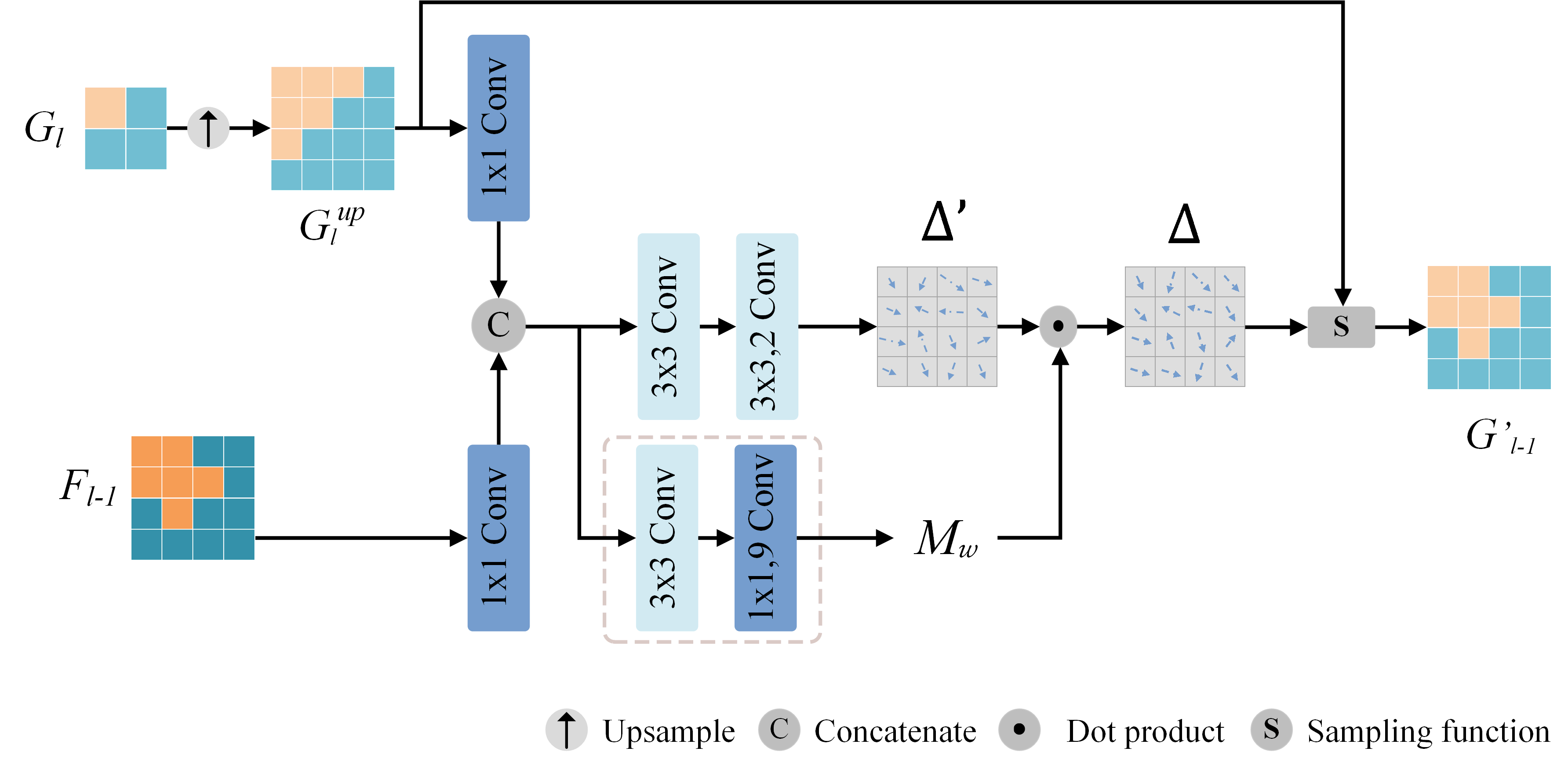}	
	\caption{Illustration of Semantic Refinement Module. SRM first takes adjacent low-resolution features and high-resolution features as inputs to predict an initial transformation offset and a weight mask $M_w$. The ultimate offset map is the result of a weighted combination of the neighborhoods on the initial offset map with the mask.}
       \label{SRM_archi}
	    \vspace{-0.3cm}
\end{figure}

\subsection{Semantic Refinement Module}

To alleviate the misalignment problem of the upsampled feature maps, the Semantic Refinement Module (SRM) is detailedly introduced in this section. 

As illustrated in Fig. \ref{SRM_archi}, the SRM takes the two feature maps $G_l$ and $F_{l-1}$ as inputs, where the resolution of $F_{l-1}$ is 2 times that of $G_l$. At first, $G_l$ is upsampled by the bilinear interpolation to the same spatial resolution as $F_{l-1}$. The upsampled feature $G_l^{up}$ and $F_{l-1}$ are compressed to the uniform channel dimension using respective $1\times1$ convolution layers. These two features are then concatenated together as $F_a$. Subsequently, the concatenated feature $F_a$ is processed through a sub-network that contains two convolutional layers, which serves to predict the initial transformation offset map $\Delta_{l-1}^\prime$. In previous works \cite{GUM,SFNet,AlignSeg}, like SFNet, the initial offset map $\Delta_{l-1}^\prime$ is used to correct the shift of pixels in $G_l^{up}$. In this paper, the entire offsets in the neighbors' area is utilized as a reference to correct the shifts of the individual pixels in this area.

To explore the influence of the neighbors’ offsets, this paper further learns a mask $M_w(k)$ to correct the offset at each pixel by taking a weighted combination over a $3 \times 3$ grid of its neighbors. The weight mask $M_w(k)$ is predicted by two convolutional layers from the concatenated features $F_a$, where $k=\left\{1,...,K\right\}$. We set $K=9$ which represents the 9 neighborhoods (including itself). The ultimate transformation offset $\Delta_{l-1}$ is computed as in Eq. (3).
\begin{equation}
	\Delta_{l-1}=\ \sum_{k=1}^{K}{( \Delta_{l-1}^\prime \cdot M_w(k))},
	\label{Delta}
\end{equation}
where $\Delta_{l-1}$ is the set of coordinate offsets $(\Delta x_i,\Delta y_i)$ for each pixel $i$ at coordinate $(x_i,y_i)$ on $G_l^{up}$.

Suppose the input feature is $G_l^{up}$ and the offset map is $\Delta_{l-1}$, differentiable image sampling function \cite{STN} is employed to generate the aligned upsampled feature $G_{l-1}^\prime$.  For each pixel $i$ on the output $G_{l-1}^\prime$, its value is computed below,
\begin{equation}
\begin{split}
		 G_{l-1}^\prime\!(i)\!\!= \!\!\sum_{m}^{H_g}\!\sum_{n}^{W_g}{G_l^{up}(m,n) max(0,\!1 \!-\! \left|x_i\!+\!\Delta x_i \!-\! m\right|)}  \\max(0,\!1 \!-\! \left|y_i\!+\!\Delta y_i  \!-\! n\right|),  
\end{split}
\end{equation}
where $G_l^{up}(m,n)$ is the value at the position $\left(m,n\right)$ on the input feature $G_l^{up}$ and $(x_i+\Delta x_i,y_i+\Delta y_i)$ is the sampling point. $ \left|x_i+\Delta x_i - m\right|$ and $\left|y_i+\Delta y_i  - n\right|$ are the distances in x-axis and y-axis between the points $(x_i+\Delta x_i,y_i+\Delta y_i)$ and the grid points $(m,n)$. The function $max(0,\cdot)$ is used to select the neighbor grid points whose distances from $(x_i+\Delta x_i,y_i+\Delta y_i)$ in x-axis and y-axis are both less than 1 to compute the value of pixel $i$. $H_g$, $W_g$ stand for the heights and widths of the input features. To further enhance the spatial semantic information, this paper aggregates the aligned upsampled feature $G_{l-1}^\prime$ and the higher-resolution feature $F_{l-1}$ of the backbone via a simple addition operation. 

 \begin{figure*}[tbp]
	\centering
	\includegraphics[width=\linewidth]{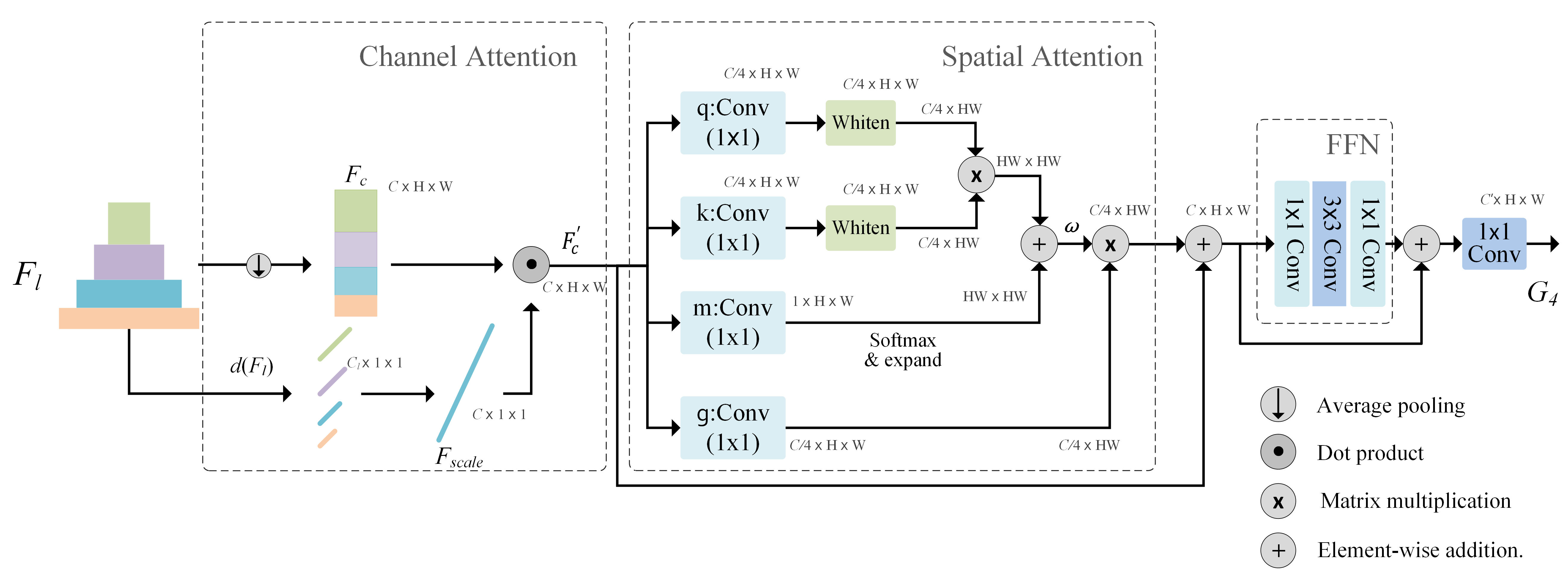}
	
	\caption{Illustration of Contextual Refinement Module. 
		The channel attention block concatenates multi-stage feature maps and models the dependencies between channels of them through the predicted scaling factor $F_{scale}$. To capture global context information in spatial dimension, the spatial attention block computes the similarity matrix $\omega$ through several transformation functions and multiplies $\omega$ with a representation of the input feature. The FFN block is employed to enhance the feature representation.}
 \label{CRM_archi}
 \vspace{-0.2cm}	
\end{figure*}
 
In this section, the semantic refinement for the upsampled feature maps is implemented by SRM. SRM first takes a low resolution feature and a high resolution feature as inputs to predict an initial transformation offset map and a weight mask, where the mask stands for the weight map for each pixel's offset. The ultimate offset map is the result of a weighted combination of the neighborhoods on the initial offset map with the mask. With the ultimate offset, SRM corrects the semantic representation of the upsampled feature maps. The effectiveness of SRM will be proven in Section IV.

\subsection{Contextual Refinement Module}
\label{}

To capture global context information along with both spatial and channel dimensions, the Contextual Refinement Module (CRM) is proposed in this section. The architecture of CRM is detailedly shown in Fig. \ref{CRM_archi}. In Fig. \ref{CRM_archi}, a series framework is constructed by a channel attention block, a spatial attention block and a Feed-Forward Network (FFN) block. The channel attention block is designed to exploit the dependencies between different channel maps from multi-stage features, while the global spatial contextual dependency between pixels are implemented by the spatial attention block. The FFN is employed to enhance the feature representation of CRM.

The channel maps can be viewed as the class-specific responses \cite{DANet}, where the different channels carry the individual semantic features. The channel attention is aimed at highlighting informative features and suppressing noisy features in all channels through the scaling factor. To enrich the semantic responses, this paper aggregates semantic maps from each of the four backbone stages to increase the channels. Given the multi-stage feature maps $\left\{F_l\right\}_{l=1,...,4} \in\mathrm{\mathbb{R}} ^ {\ C_l \times \frac{H}{2^{l+1}} \times \frac{W}{2^{l+1}}  } $ as inputs, we concatenate the multi-stage feature maps together and predict the scaling factor $F_{scale}$ to re-calibrate the concatenated feature maps. Specifically, the scaling factor $F_{scale}$ is predicted by employing the global pooling operation and the light MLP blocks. In this work, we first predict four preliminary scaling factors from the four stage feature maps. These factors are then fused together to predict the final scaling factor $F_{scale}$.
This process is expressed in Eq. (5). 
\begin{equation}
	F_{scale} = \delta(MLP_{fuse}(cat(d_1(F_1),...,d_l(F_l)))), 
\end{equation}
where $d_i(F_i)_{i=1,\cdots,l}=MLP_{l} (pooling_{g}(F_i))$, and the $pooling_{g}(\cdot)$ denotes the global pooling operator. $MLP_{fuse}$ and $MLP_{l}$ are the light MLP blocks that include two fully-connected layers around the ReLU function. $\delta(\cdot)$ represents the sigmoid activation function. After that, the re-calibrated features ${F_c}^{'}$ generated by multiplying the scaling factor $F_{scale}$ and the concatenated feature $F_c$, is written as
\begin{equation} 
	{F_c}^{'} = F_{scale} \cdot F_c,
\end{equation}
where $F_c$ is calculated by pooling the backbone's four stage features, which are of different scales, to the uniform size $\frac{H}{32}\times\frac{W}{32}$ and then concatenating them together. The concatenation process is formulated as
\begin{equation} 
	\!\!\! F_c\!\!=\!cat(pooling_{a}(F_1),pooling_{a}(F_2) ,pooling_{a}(F_3) , F_4),
\end{equation} 
where $cat(\cdot)$ and $pooling_{a}(\cdot)$ indicate the concatenation operator and the average pooling operator.

Within the spatial dimension, the spatial attention block, based on Disentangled Non-Local (DNL) \cite{DNL}, investigates the correlation between pixels in the re-calibrate concatenated feature map ${F_c}^{'}$. The proposed spatial attention block adaptively assigns weights to different positions by analyzing the similarity between pixels, thereby capturing contexts based on their contribution. For $x_i$ and $x_j$ in ${F_c}^{'}$, the similarity $\omega\left(x_i,x_j\right)$ between them is calculated by Eq. (8).
\begin{equation} 
	\omega\left(x_i,x_j\right)=\sigma\left({(q_i-\mu_q)}^T(k_j-\mu_k)\right)+\sigma(m_j),
\end{equation}
where $q_i = W_qx_i$, $k_j=W_kx_j$, and $m_j=W_mx_j$ denote embeddings of corresponding pixel. The $W_q$, $W_k$\ and $W_m$ are the weights that need to be learned. $\mu_q$\ and $\mu_k$\ are the averages of $q_i$ and $k_j$ over all pixels, computed as $\frac{1}{\left| \Omega \right|} \sum_{i\in\mathrm{\Omega}}{q_i}$ and $\frac{1}{\left| \Omega \right|} \sum_{j\in\mathrm{\Omega}}{k_j}$ respectively. $\mathrm{\Omega}$ represents the collection of all pixels. $\sigma(\cdot)$ stands for the SoftMax function. 
For pixel $i$ with similarity function $\omega\left(x_i,x_j\right)$ from pixel $j$, the output feature $y_i$ of the spatial attention block is calculated as
\begin{equation} 
	y_i=\sum_{j\in\mathrm{\Omega}}{\omega(x_i,x_j)g(x_j)}, 
\end{equation} 
where $g\left(x_j\right)$ represents the transformation for $x_j$. Fig. \ref{CRM_archi} illustrates the details of the spatial attention block. Specifically, the transformation functions for $q$, $k$, $m$ and $g\left(x\right)$ are executed using a $1\times1$ convolution layer respectively. The processes of $(q_i-\mu_q)$ and $(k_j-\mu_k)$ are marked as ``Whiten" in Fig. \ref{CRM_archi}. 

The FFN block is employed to improve the feature representation ability of CRM. Its architecture encompasses two $1\times1$ convolution layers, with a depth-wise convolution layer inserted between them.

In this section, CRM enhances the capacity to capture context information. Based on the aggregation of semantic maps from multiple stages in the backbone, the channel attention block highlights the contribution of the high-dependency pixels from different channels to the semantic label alignment. On the other hand, the spatial attention block exploits the dependencies between pixels in the spatial dimension to capture global context information. With the refined features by CRM, the segmentation performance is drastically improved, as shown in Section IV.

\subsection{Training Loss}
This section presents the loss function utilized for our semantic segmentation network.  It comprises both cross-entropy loss and contrastive loss.

The cross-entropy loss is applied to supervise the segmentation prediction results of the network. It is formulated as
\begin{equation}
	L^{ce}=-\sum_{i=1}^{n}{y_ilog{(p}_i)},
\end{equation} 
where, $y_i$ represents the ground truth for each given pixel $i$ and $p_i$ stands for the predicted segmentation probability.  
 
To increase the intra-class compactness and the inter-class separability, the contrastive loss \cite{closs} is adopted as an auxiliary loss function. Its purpose is to bring pixels belonging to the same class closer together and push those from different classes further apart in the embedding space \cite{wang2024mix}. It is defined in Eq. (11).
\begin{equation}
	 L_i^{cl}  = \frac{1}{\left|p_i\right|}\   \sum_{i^+\in P_i} {-log\  \frac{e^{(i\cdot i^+/\tau)}}{e^{(i\cdot i^+/\tau)} + \sum_{i^-\in \! N_i} {e^{(i\cdot i^-/\tau)}}}} ,
	 \label{contrastiveloss}
\end{equation}
where $\tau>0$ is a hyper-parameter. For each pixel $i$ in the embedding space, its variations $i^+$ and $i^-$ represent pixels belonging to the same class and the different classes, separately. A $1\times1$ convolution layer with 256 channels, followed by a normalization layer, is employed as the embedding head to project the pixels into the embedding space. This embedding head is used only during the training phase, ensuring that there is no additional computational burden during inference.

Overall, the total loss function is written in Eq. (12).
\begin{equation}
	L=L^{ce}+\lambda\ L^{cl},
\end{equation}
where parameter $\lambda$ stands for the coefficient of $L^{cl}$ to balance the two losses.

\section{Experiments}
\label{sec:Experiments}
\subsection{Datasets}
\label{ssec:Datasets} 
We evaluate the effectiveness of the proposed method using three semantic segmentation datasets: Cityscapes \cite{Cityscapes}, Bdd100K \cite{BDD100K}, and ADE20K \cite{ADE20K}. ADE20K contains 150 semantic categories for scene parsing, while Cityscapes and Bdd100K are urban street scene datasets with 19 semantic categories. Compared to Cityscapes, Bdd100K is a newer and more challenging dataset with geographic, environmental, and weather diversity. Cityscapes includes 5,000 finely annotated images with high resolution ($2048 \times 1024$), which are divided into 2,975 for training, 500 for validation, and 1,525 for testing. Bdd100K comprises 7,000 training images, 1,000 validation images, and 2,000 testing images, each with a resolution of $1280\times720$. The ADE20K dataset includes 20,210 images for training, 2,000 for validation, and 3,000 for testing. The images in the ADE20K dataset are from various scenes and have more scale variation. 
Follow the same setting of existing methods \cite{Shao2024CoTCT, Lateef2024ATC}, this paper chooses the initial division of the datasets to train and evaluate the proposed model.

\subsection{Implementation Details}
\label{ssec:Implementation Details}
We implemented the proposed network using the PyTorch. In our experiment, data augmentation is conducted through random horizontal, random resizing, and random cropping. The random resizing is applied within a scale range of 0.5 to 2.0. For random cropping, the size is set to $1024 \times 1024$ for Cityscapes, $720\times720$ for Bdd100K, and $512\times512$ for ADE20K. During the training phase, we utilize the Adam optimizer and the poly learning rate policy. The initial learning rate is set at 0.00006 for Cityscapes and 0.00012 for Bdd100K and ADE20K. For the Cityscapes dataset, we set the batch size to 16, while for both Bdd100K and ADE20K, the batch size is configured at 32. For all three datasets, the training epoch is set to 600. Due to the hardware resource limitation, this paper only implements the large model training on Cityscapes, where the pretrained MSCAN-L \cite{SegNeXt} on ImageNet \cite{ImageNet} is utilized as the backbone. To confirm the generality and effectiveness of the proposed modules, we also train other two light-weight models on all three datasets. In the light-weight models, the pretrained MSCAN-S \cite{SegNeXt} and VAN-S \cite{van} are employed.

In the DNL block, the channel dimension of $q$ and $k$ is reduced to $C/4$, where $C$ represents the channel dimension of the input features. Regarding the loss function parameters, we configure $\lambda$ at 1 and $\tau$ at 0.1 respectively. In computing the contrastive loss (cf. Eq. (\ref{contrastiveloss})), we follow the segmentation-aware hard anchor sampling strategy in \cite{closs} and sample 1024 anchors for each mini-batch.

In our experiment, the segmentation head includes two $1\times1$ convolution layers, between which a BatchNorm layer and a ReLU activation layer are integrated.

\begin{table}[tbp]
	\centering
	\renewcommand\arraystretch{1.3}
	\caption{Evaluation of the proposed modules on the Cityscapes val set. The mIoU here is evaluated with single-scale inference. The “MF” and “$F_4$” stand for the multi-stage feature maps and the single last stage feature map of the backbone. “\textnormal{Bilinear}” denotes bilinear upsampling.}
	\setlength{\tabcolsep}{1mm}{
		\begin{tabular}{ccc:cc:c:c}
			\toprule
			& MF & $F_4$ & SRM  & Bilinear   & CRM  & mIoU(\%) $\uparrow$    \\   \hline
			
			Baseline     &    & \checkmark  &   & \checkmark & &  81.3     \\
			Baseline+SRM  &    & \checkmark  & \checkmark  & &  &  82.0 (+0.7)    \\
			Baseline+CRM &  \checkmark &    &  & \checkmark & \checkmark&  81.8 (+0.6)    \\  \midrule
			
			Ours         &  \checkmark &    & \checkmark & & \checkmark & 82.5 (+1.2)   \\ 
			\bottomrule
	\end{tabular} }
    \label{base_ablation}
    \vspace{-0.2cm}
\end{table}

\subsection{\bf Ablation study} 
We conduct ablation experiments to evaluate the influence of the proposed modules, including SRM and CRM. The feature pyramid network architecture with MSCAN-S is taken as the {\bf baseline}. In the baseline, the feature map $F_4$ from the stage 4 of MSCAN-S is used as input to the decoder. The results on the Cityscapes val set are presented in TABLE \ref{base_ablation}. The number of FLoating-point OPerations (FLOPs) and the parameters of the network are also presented as the reference for measuring the size of the models. The mean of class-wise Intersection over Union (mIoU) is employed to evaluate the performance of various methods. By replacing bilinear upsampling with SRM, ``Baseline+SRM" improves 0.7\% mIoU than the Baseline.  Adding CRM, denoted as ``Baseline+CRM", brings 0.6\% mIoU improvements. Benefiting from the joint utilization of CRM and SRM, the proposed method brings 1.2\% mIoU improvements. The improvements demonstrate the effectiveness of the proposed modules. 
\subsection {\bf SRM vs. Other alignment modules}We compare SRM with other alignment modules on three datasets. TABLE \ref{align_ablation} shows the comparing results with the bilinear upsampling, Aligned Feature Aggregation (AlignFA) \cite{AlignSeg}, the Flow Alignment Module (FAM) \cite{SFNet} and our SRM. All methods listed in TABLE \ref{align_ablation} employ CRM for context modeling. In comparison to FAM and AlignFA, SRM improves mIoU with a small extra calculation burden.

\begin{table}[tbp]
	\centering
	\renewcommand\arraystretch{1.3}
	\caption{Comparisons with different alignment modules on Cityscapes val, Bdd100K and ADE20K set. The GFLOPs are calculated on Cityscapes with the image resolution is set to $2048\times1024$. ``G” and  ``P” stand for GFLOPs and Params respectively.}
\begin{tabular}{cccccc}
\hline
\multirow{2}{*}{Method} & \multicolumn{3}{c}{mIoU(\%) $\uparrow$} & \multirow{2}{*}{G $\downarrow$} & \multirow{2}{*}{P(M)$\downarrow$} \\ \cline{2-4}
                       & Cityscapes & Bdd100K & ADE20K &        &       \\ \hline
Bilinear               &    81.8      &   64.3     &   44.4    & 122.15 & 15.21 \\
FAM \cite{SFNet}       &     82.1    &   65.1      &   44.8     & 124.31 & 15.45 \\
AlignFA \cite{AlignSeg}   &   82.0       &  65.3      &   44.9     & 134.31 & 15.6  \\
SRM &   {\bf82.5}      &    {\bf65.9}     &   {\bf45.2}     & 137.83 & 15.49 \\ \hline
\end{tabular}
\label{align_ablation}
\vspace{-0.2cm}
\end{table}
\begin{table}[tbp ]
	\centering
	\renewcommand\arraystretch{1.3}
	\caption{Comparisons with different context modeling modules on Cityscapes val, Bdd100K and ADE20K set. The GFLOPs are calculated on Cityscapes with the image resolution is set to $2048\times1024$. ``G” and  ``P” stand for GFLOPs and Params respectively.}
\begin{tabular}{cccccc}
\hline
\multirow{2}{*}{Method} & \multicolumn{3}{c}{mIoU(\%) $\uparrow$} & \multirow{2}{*}{G $\downarrow$} & \multirow{2}{*}{P(M)$\downarrow$} \\ \cline{2-4}
                &  Cityscapes & Bdd100K & ADE20K &        &       \\ \hline
Baseline+MF     &    82.0     &   64.7     &   44.5      & 135.13 & 13.99  \\
PPM\cite{PPM}      &  82.1        &      65.2    &   44.9     & 135.28 & 14.58 \\
DAPPM\cite{DDRNet}  &  82.0       &   65.2     &   44.6    & 136.89 & 15.33  \\
CCAM\cite{CCNet}      &  82.0       &  65.5       &   44.6   & 137.82 & 15.3 \\
DAM\cite{DANet}  &    82.2     &    65.6     &   44.7      &138.69 & 15.73  \\
CRM &   {\bf82.5}      &   {\bf65.9}     &  {\bf45.2}      & 137.83 & 15.49 \\ \hline
\end{tabular}
\label{context_ablation}
\vspace{-0.3cm}
\end{table}

\begin{table*}[hbp]
	\centering
	\caption{Comparison on Cityscapes val and test set with latest state-of-the-art models. The mIoU here is evaluated with multi-scale inference. ``Size" refers to the image resolution used to calculate GFLOPs. † denotes that the backbone is pre-trained on ImageNet-22K. The test set result is trained on the combined ``train+val'' set.}
	\renewcommand\arraystretch{1.3}
	\setlength{\tabcolsep}{4mm}{
		\begin{tabular}{cccccccc}
			\hline
			\multirow{2}{*}{Method} &
			\multirow{2}{*}{Publication} &
			\multirow{2}{*}{Backbone} &
			\multicolumn{2}{c}{mIoU (\%) $\uparrow$} &
			\multirow{2}{*}{Size} &
			\multirow{2}{*}{GFLOPs $\downarrow$ } &
			\multirow{2}{*}{Params (M) $\downarrow$ } \\ \cline{4-5}
			&             &             & val   & test  &           &        &        \\ \hline
			DANet \cite{DANet}     & CVPR2018    & ResNet-101  & 81.5 & 81.5 &    -     &  -   &   -    \\
			CCNet \cite{CCNet}     & ICCV2019    & ResNet-101  & 81.3  & 81.4  &    -     &   -   &  -    \\
			SFNet \cite{SFNet}     & ECCV2020    & ResNet-101  & -   & 81.8  & $1024\times1024$  & 417.5  & 50.3  \\
			HRNetV2 \cite{HRNet}   & TPAMI2021   & HRNetV2-W48 & -    & 81.6  & $2048\times1024$ & 696.2  & 65.9   \\
			HRNetV2+OCR \cite{HRNet}      & TPAMI2021   & HRNetV2-W48 &  -   & 82.5  & $2048\times1024$  & 1206.3 &  70.3  \\
			NRD \cite{NRD}      & NeurIPS2021 & ResNet-101  & -    & 80.5  & $2048\times1024$ & 390.0  &   -   \\
			AlignSeg \cite{AlignSeg}  & TPAMI2021   & ResNet-101  & 82.4  & 82.6  &   -      &   -   &   -   \\
			SegFormer \cite{SegFormer} & NeurIPS2021 & MiT-B5      & 84.0  & 82.2  & $2048\times1024$ & 1447.6 & 84.7   \\
			IFA \cite{hu2022learning}       & ECCV2022    & ResNet-101  & -   & 82.0  &  $1024\times1024$ & 281.4  & 64.3   \\ 
            SRRNet \cite{SRRNet}  & TMM2022   & ResNet-101  & -  & 82.3  &   -      &   -   &   -   \\
			SegNeXt \cite{SegNeXt}  & NeurIPS2022 & MSCAN-L     & 83.9  &  -   & $2048\times1024$ & 577.5  & 48.9   \\
            CANet \cite{CANet}  & TMM2023 & ResNet-101     & -  & 82.1   & -  & -   & 65.2  \\
            OneFormer \cite{OneFormer}  & CVPR2023 & Swin-L     & 84.4  &  -   & $1024\times512$  & 543   & 219  \\
            CFT \cite{CFT}  & Arxiv2023 & Swin-L†     & 84.4  &   83.3    & $2048\times1024$ & 2130 & 200   \\
			DDP \cite{DDP}  & ICCV2023   & Swin-B      & 83.4 &   -  & $2048\times1024$ & 1357   & 99     \\ 
            HFGD \cite{Huang2023HighlevelFG}  & TCSVT2024  & ConvNeXt-L   & 84.0 &   83.3 & - & -   & -     \\ 
            CoT \cite{Shao2024CoTCT}  & TNNLS2024   & Swin-L   & 84.2 &   -  & -   & -   & -     \\  \hline
			Ours      &             & MSCAN-L     &  {\bf 84.5}  &   {\bf 83.8}    &  $2048\times1024$  &    411.2    &   46.7     \\ \hline
	\end{tabular}}
    \label{ml_city}
\end{table*}

\subsection{\bf CRM vs. Other context modeling modules} We further explore the influence of CRM by comparing it with other methods. The proposed CRM is compared with other context modeling modules, including Pyramid Pooling Module (PPM) \cite{PPM}, Deep Aggregation Pyramid Pooling Module (DAPPM) \cite{DDRNet}, Criss-Cross Attention Module(CCAM) \cite{CCNet}, and Dual Attention Module (DAM) \cite{DANet}. To ensure a fair comparison, the input for the above context modeling methods is set as the multi-stage features. TABLE \ref{context_ablation} presents the results of the comparison on three datasets. The “Baseline+MF” in TABLE \ref{context_ablation} means the baseline with the multi-stage features as input. All methods in TABLE \ref{context_ablation} use SRM in the decoder phase. One thing to note here is that we remove the last convolutional layer in each branch of DAM to ensure similar computational cost and parameters as other methods. From TABLE \ref{context_ablation}, it is shown that CRM achieves better mIoU performance than the other context modeling modules by using the similar GFLOPs and Params.

\begin{table*}[htp]
	\centering
	\caption{ Comparison on Cityscapes val and test set with the latest light-weight models. The mIoU here is evaluated with single-scale inference. ``Size" refers to the image resolution used to calculate GFLOPs. The test set result is trained on the combined ``train+val'' set.}
	\renewcommand\arraystretch{1.3} 
	\setlength{\tabcolsep}{4mm}{
		\begin{tabular}{@{}ccccccccc@{}}
			\toprule
			\multirow{2}{*}{Method} &
			\multirow{2}{*}{Publication} &
			\multirow{2}{*}{Backbone} &
			\multicolumn{3}{c}{mIoU (\%) $\uparrow$} &
			\multirow{2}{*}{Size} &
			\multirow{2}{*}{ GFLOPs $\downarrow$ } &
			\multirow{2}{*}{Params (M) $\downarrow$ } \\ \cmidrule(lr){4-6}
			&             &               & \multicolumn{2}{c}{val}   & test     &           &        &        \\ \midrule
			SFNet \cite{SFNet}      & ECCV2020    & ResNet-18  & -    & \multicolumn{2}{c}{78.9} & $2048\times1024$ & 243.87 & 12.87  \\
			SegFormer \cite{SegFormer}  & NeurIPS2021 & MiT-B1  & 78.5   & \multicolumn{2}{c}{-}    & $2048\times1024$ & 243.7  & 13.7  \\
			BiAlignNet \cite{BiAlignNet} & ICIP2021    & DFNet2 & 78.7   & \multicolumn{2}{c}{77.1} & $2048\times1024$ & 108.73 & 19.2  \\	 
			SFANet \cite{SFANet}     & TCSVT2022  & ResNet-18 & -   & \multicolumn{2}{c}{78.1} & $2048\times1024$ & 99.6   & 14.6  \\
			IFA \cite{hu2022learning}        & ECCV2022    & ResNet-50     & 78.0         & \multicolumn{2}{c}{-}     & $1024\times1024$ & 186.9  & 27.8   \\
			RTFormer \cite{RTFormer}   & NeurIPS2022 & RTFormer-Base & 79.3         & \multicolumn{2}{c}{-}    & $2048\times1024$ & -      & 16.8   \\
			\multirow{2}{*}{DDRNet \cite{DDRNet}} &
			\multirow{2}{*}{TITS2022} &
			DDRNet-23 &
			79.5 &
			\multicolumn{2}{c}{79.4} &
			$2048\times1024$ &
			143.1 &
			20.1   \\
			&  & DDRNet-39 & 80.4  & \multicolumn{2}{c}{80.4} & $2048\times1024$ & 281.2  & 32.3  \\ 
			SegNeXt \cite{SegNeXt} & NeurIPS2022    & MSCAN-S & 81.3   & \multicolumn{2}{c}{-} & $2048\times1024$ & 124.6 & 13.9 \\
			\multirow{2}{*}{PRSeg \cite{PRSeg} } &
			\multirow{2}{*}{TCSVT2023} &
			PRSeg-S &
			79.5 &
			\multicolumn{2}{c}{-} &
			$768\times768$ &
			246 &
			26  \\
		    & & PRSeg-M  & 79.2   & \multicolumn{2}{c}{-} & $768\times768$ & 65  & 30  \\  
		    PIDNet \cite{PIDNet}     & CVPR2023   & PIDNet-M      & 79.9         & \multicolumn{2}{c}{79.8} & $2048\times1024$ & 178.1  & 28.5   \\ 
		    FRMSeg \cite{wang2023feature}     & ICIP2023   & VAN-S      & 80.8         & \multicolumn{2}{c}{80.4} & $2048\times1024$ & 214.82  & 16.48   \\
      		SCTNet \cite{sctnet}  & AAAI2024  & SCTNet-B-Seg100   & 80.5       & \multicolumn{2}{c}{-} &  $2048\times1024$ & -  & 17.4   \\ 
            HSNet \cite{HSNet}  &TSMC-S2024  &STDC2    & -         & \multicolumn{2}{c}{ 80.0 } & $1024\times512$ & 33  & 16.4 \\ \midrule
			\multirow{2}{*}{Ours} &
			\multirow{2}{*}{ } &
			VAN-S &
			82.1 &
			\multicolumn{2}{c}{\bf 81.5} &
			$2048\times1024$ &
			135.1  &
			15.4   \\
			&  & MSCAN-S & {\bf 82.5}   & \multicolumn{2}{c}{\bf 81.5} & $2048\times1024$ & 137.9  & 15.5   \\  \bottomrule
	\end{tabular}}
	\label{ms_city}
    \vspace{-0.3cm}
\end{table*}

\begin{figure*}[hbp]
	\centering
	\includegraphics[width=1\linewidth]{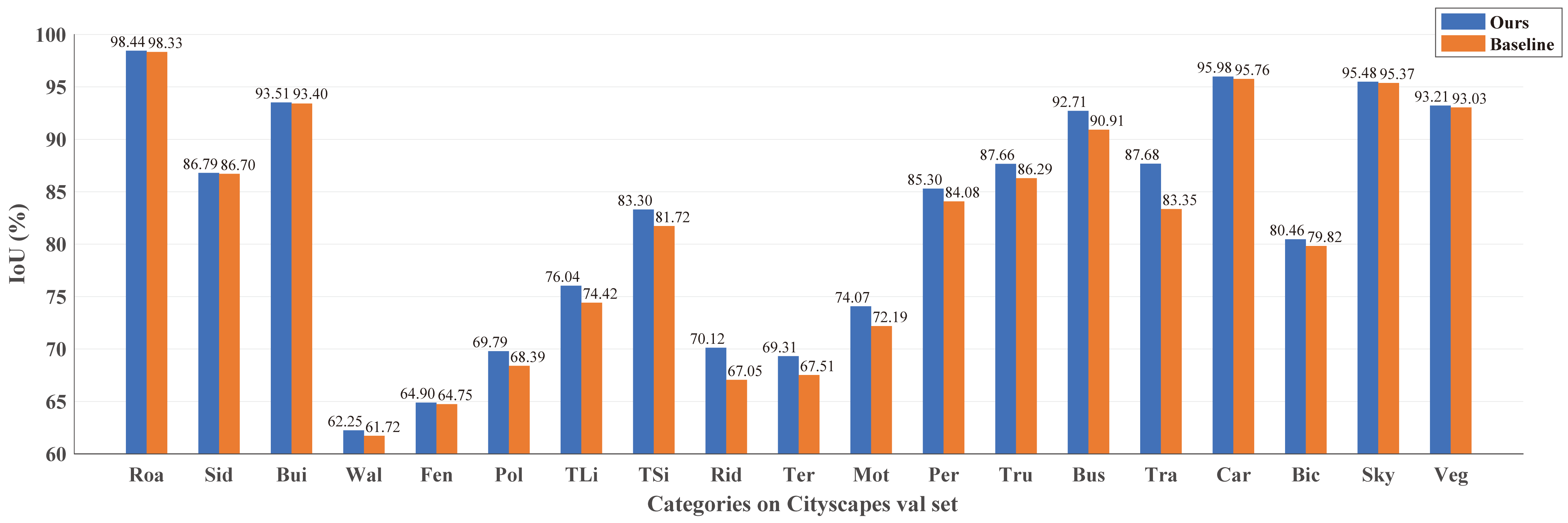}
	
	\caption{Comparison of per category results on Cityscapes val set. The vertical axis shows IoU (\%) for each category, while the horizontal axis indicates the categories within the dataset. List of categories (from left to right): road, sidewalk, building, wall, fence, pole, trafficlight, trafficsign, rider, terrain, motorcycle, person, truck, bus, train, car, bicycle, sky, vegetation.}

	\label{category_results}
\end{figure*}

\subsection{Experiment on Cityscapes}

\subsubsection{\bf Comparison with State-Of-The-Art (SOTA)}
TABLE \ref{ml_city} shows the comparison of results on Cityscapes dataset with state-of-the-art methods. The compared methods include several CNN based methods 
DANet \cite{DANet}, CCNet \cite{CCNet}, SFNet \cite{SFNet}, HRNet \cite{HRNet}, NRD \cite{NRD}, AlignSeg \cite{AlignSeg}, IFA \cite{hu2022learning}, SRRNet \cite{SRRNet}, CANet \cite{CANet}, SegNeXt \cite{SegNeXt}, HFGD \cite{Huang2023HighlevelFG}, and Vision Transformer based methods SegFormer \cite{SegFormer}, OneFormer \cite{OneFormer}, CFT \cite{CFT}, DDP \cite{DDP}, CoT \cite{Shao2024CoTCT}. In this subsection, multi-scale inference is performed to ensure a fair comparison with other methods. 

From TABLE \ref{ml_city}, it is shown that the proposed method attains 84.5\% mIoU on the Cityscapes val set and 83.8\% mIoU on the Cityscapes test set, achieving the best performance with a limited  GFLOPs and Params. In comparison with the previous SOAT methods OneFormer \cite{OneFormer} and CFT \cite{CFT} on the Cityscapes val set, the proposed method obtains a slight performance gain in mIoU, but with only about $1/5$ of computational cost. Compared to SegNeXt-L \cite{SegNeXt} with similar computational cost and  parameters, the proposed method with the same MSCAN-L backbone achieves a 0.6\% improvement in mIoU on the Cityscapes val set while requiring only 71\% of the computational cost.

\subsubsection{\bf Comparison with light-weight models} 
Since the proposed SRM and CRM are both lightweight modules, they are also suitable for lightweight semantic segmentation networks. TABLE \ref{ms_city} shows the comparison results on Cityscapes dataset with several light-weight networks, including the methods based on multi-branch structure, BiAlignNet \cite{BiAlignNet}, RTFormer \cite{RTFormer}, DDRNet \cite{DDRNet}, PIDNet \cite{PIDNet}, and the methods based on encoder-decoder structure, SFNet \cite{SFNet}, SegFormer \cite{SegFormer}, NRD \cite{NRD}, SFANet \cite{SFANet}, IFA \cite{hu2022learning}, PRSeg \cite{PRSeg}, FRMSeg \cite{wang2023feature}, SCTNet \cite{sctnet}, HSNet \cite{HSNet}. The result of mIoU is evaluated with single-scale inference. To confirm the generality of the proposed method, we employ two light-weight backbones, namely MSCAN-S and VAN-S, for the model training. It is observed from TABLE \ref{ms_city} that the proposed method achieves better performance than the other light-weight methods no matter which of the two backbones is employed. The proposed method, when implemented with VAN-S on the Cityscapes test set, achieves 81.5\% mIoU with 135.1 GFLOPs for a $2048 \times 1024$ input. Compared to the previous SOTA lightweight network DDRNet-39 on the Cityscapes test set, the proposed method achieves a 1.1\% mIoU increase in mIoU with only 49\% of computational cost and 48\% of parameters. Compared with our previous preliminary work \cite{wang2023feature}, we further improve the 1.1\% mIoU on the Cityscapes test set and 1.3\% mIoU on the Cityscapes val set with the same backbone VAN-S. With the similar GFLOPs and Params, there is a 1.2\% mIoU increase for the proposed method when compared to SegNeXt-S \cite{SegNeXt}.

\subsubsection{\bf Results of each category} 
This paper also shows the IoU performance for each category on Cityscapes val set. Fig. \ref{category_results} displays the comparative performance between the proposed method and the baseline. From Fig. \ref{category_results}, it is observed that the proposed method outperforms the baseline in almost all categories, especially in the categories ``rider", ``terrain", ``motorcycle" etc.

\begin{figure*}[htb]
	\centering
	\captionsetup[subfloat]{labelsep=none,format=plain,labelformat=empty}
	\subfloat{\includegraphics[width=4.4cm]{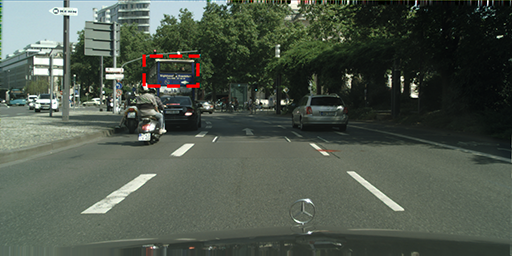}%
	}
	\hspace{0cm}
	\subfloat{\includegraphics[width=4.4cm]{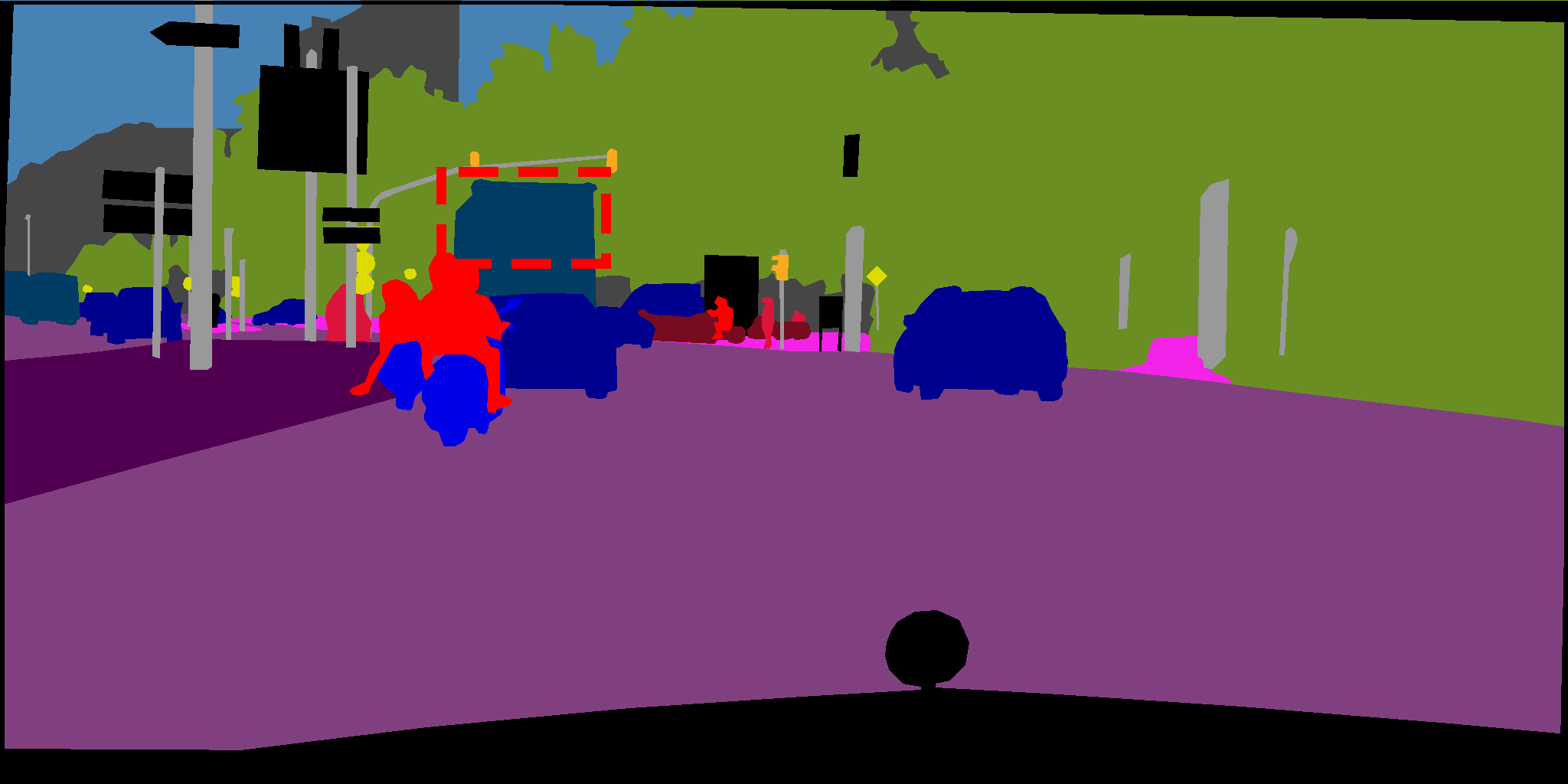}%
	}
	\hspace{0cm}
	\subfloat{\includegraphics[width=4.4cm]{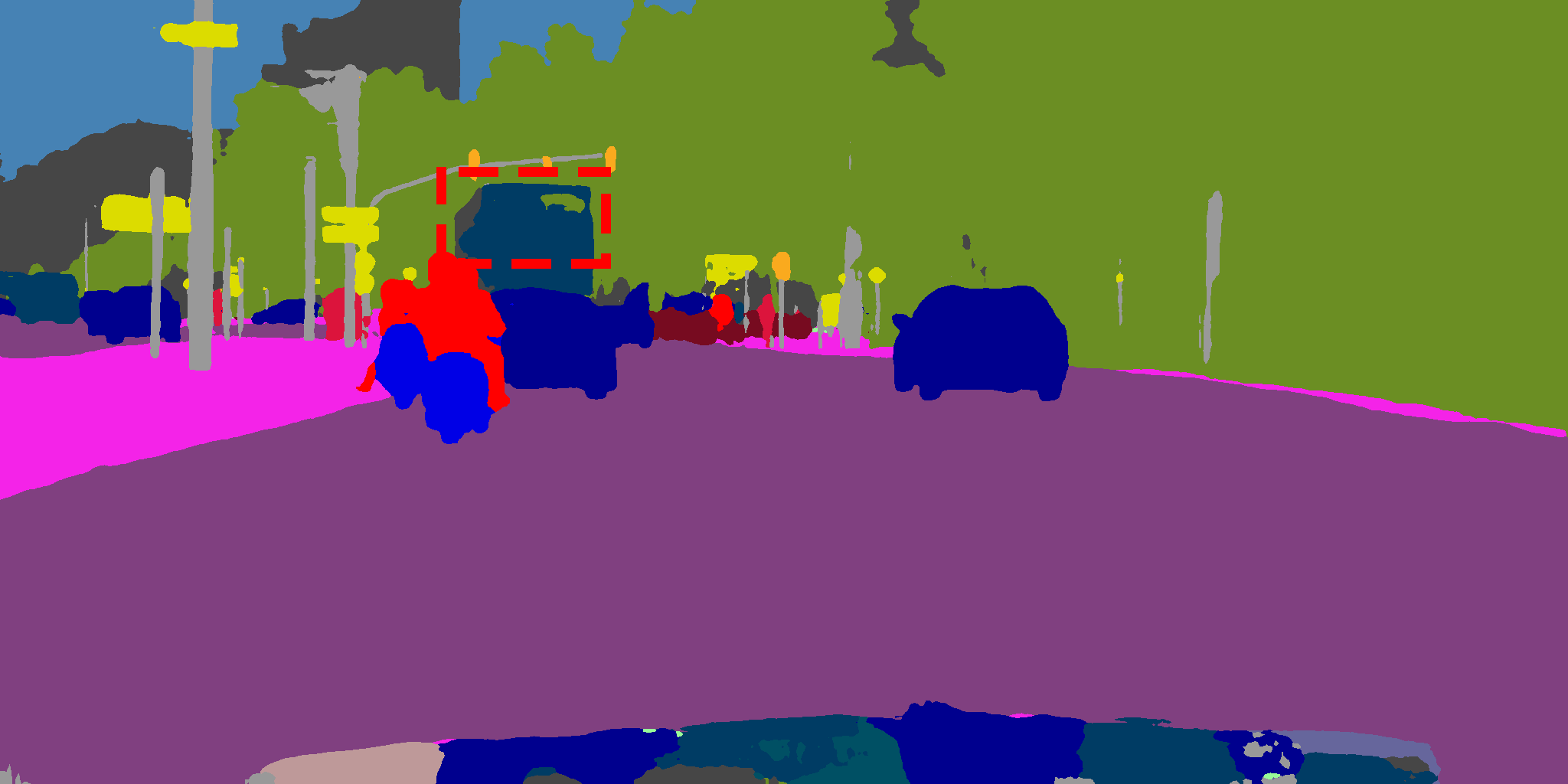}%
	}
	\hspace{0cm}
	\subfloat{\includegraphics[width=4.4cm]{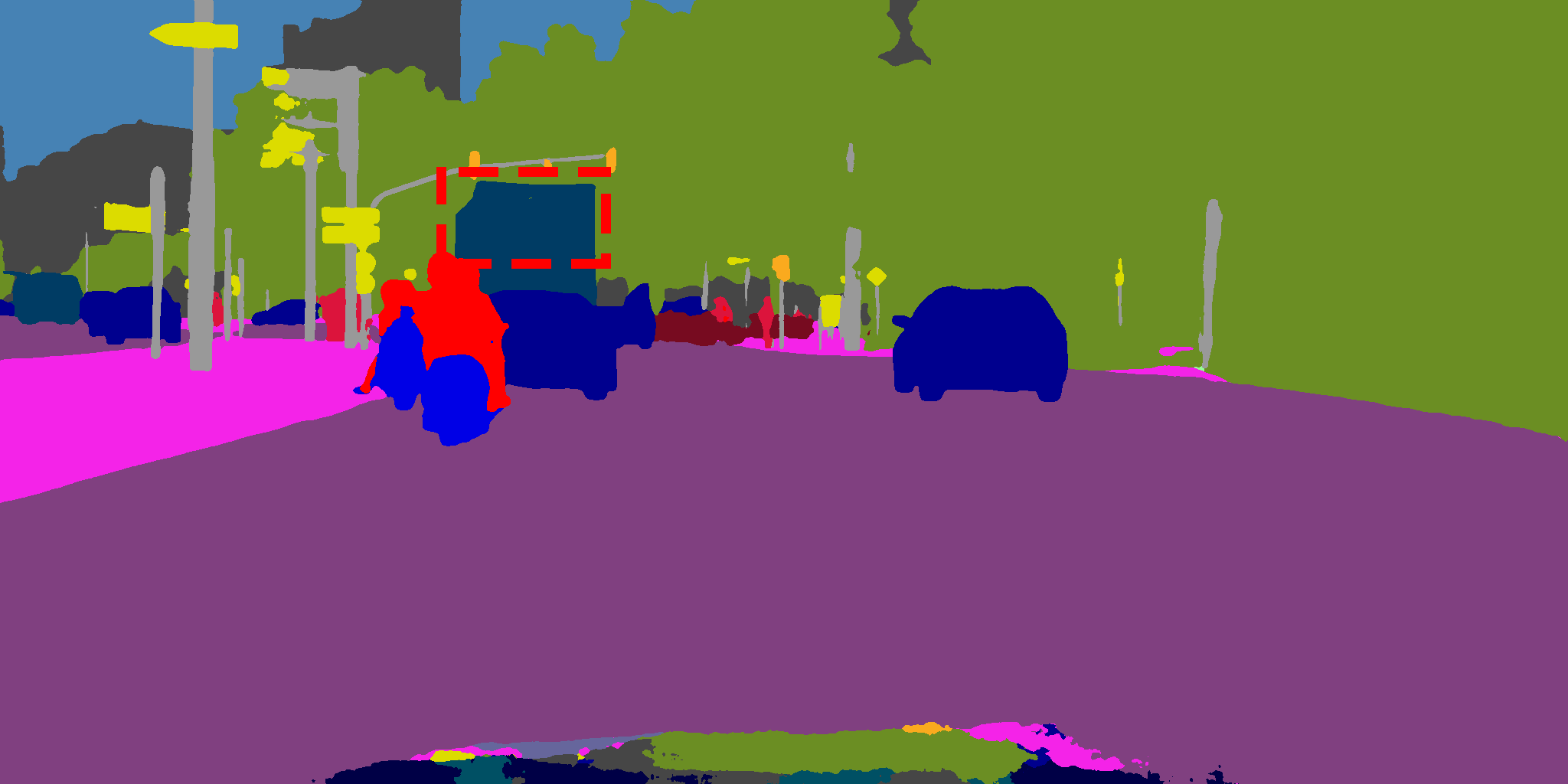}%
	}
	\vspace{0.15cm}
	\subfloat{\includegraphics[width=4.4cm]{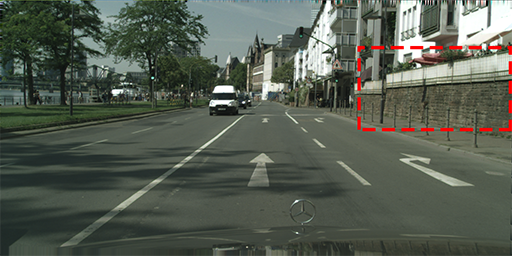}%
	}
	\hspace{0cm}
	\subfloat{\includegraphics[width=4.4cm]{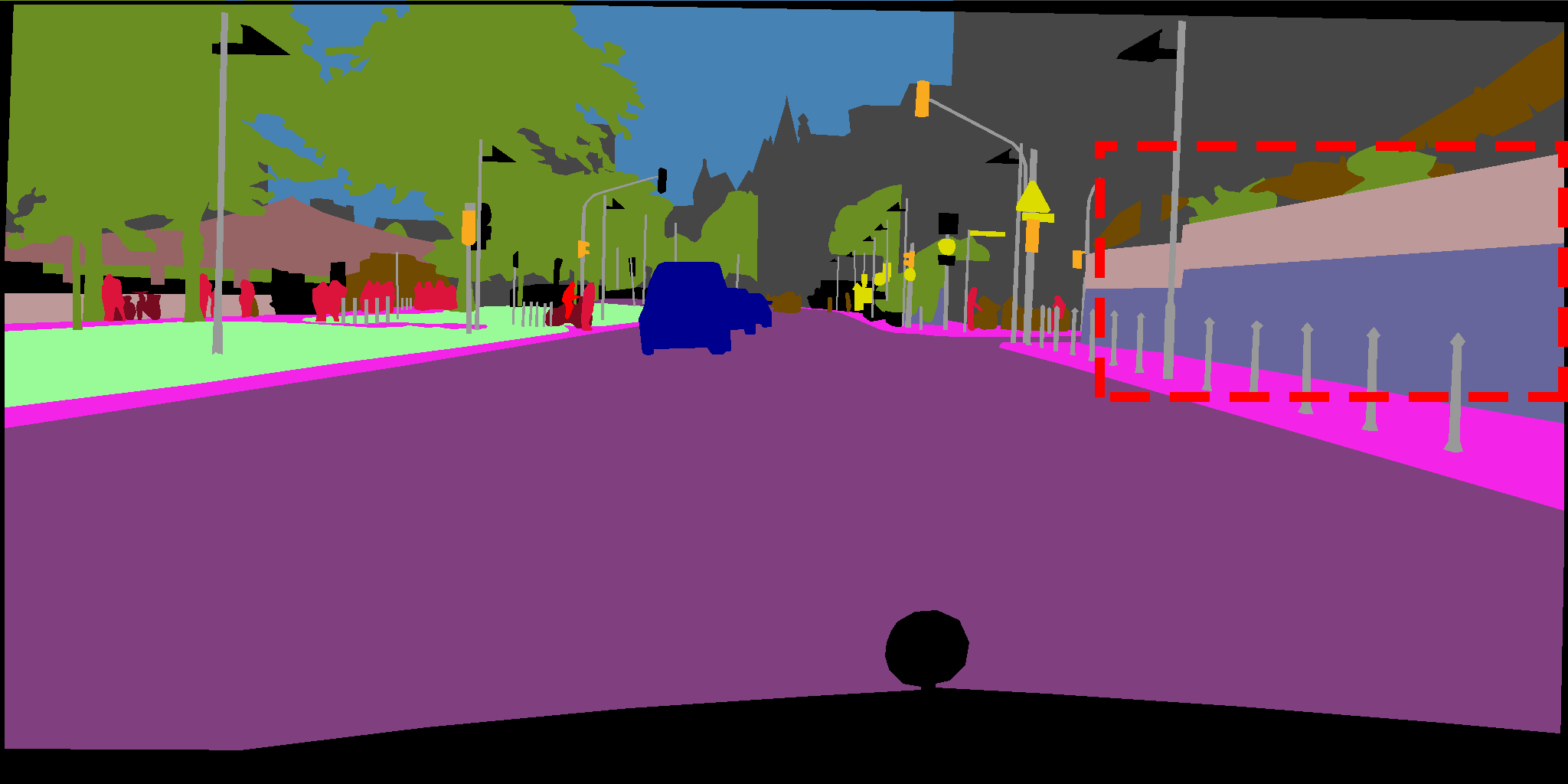}%
	}
	\hspace{0cm}
	\subfloat{\includegraphics[width=4.4cm]{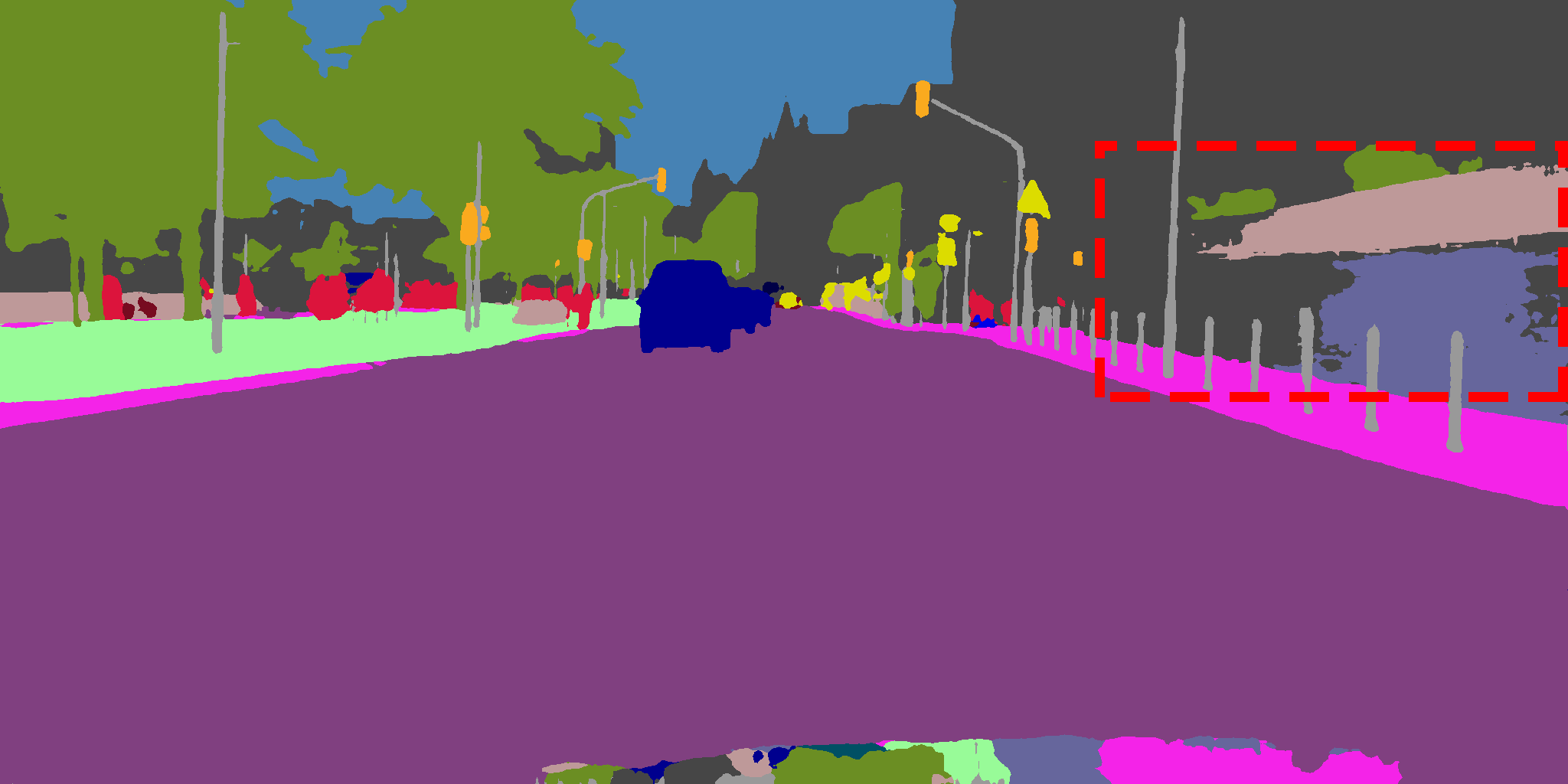}%
	}
	\hspace{0cm}
	\subfloat{\includegraphics[width=4.4cm]{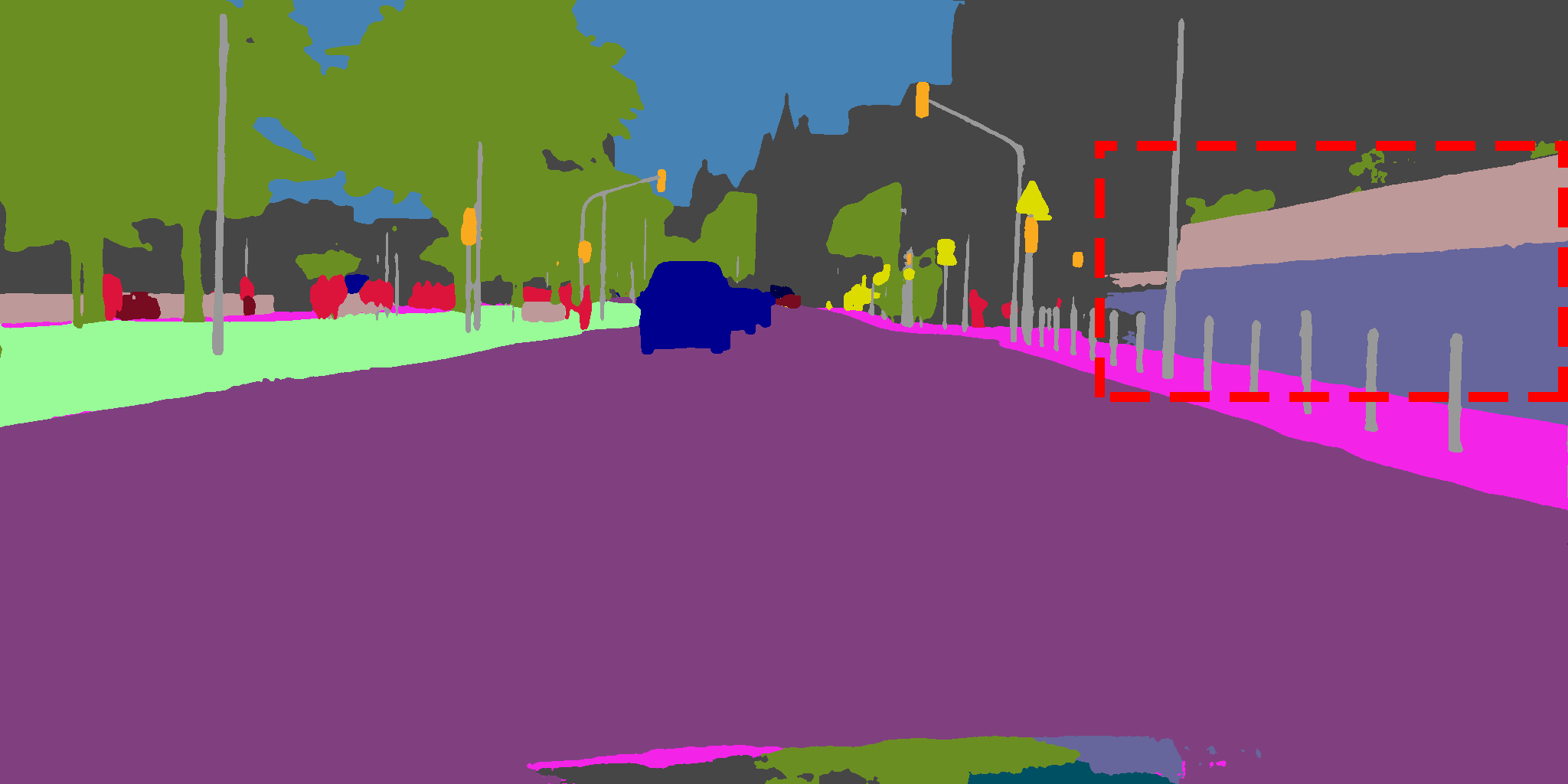}%
	}
	\vspace{0.15cm}
	\subfloat{\includegraphics[width=4.4cm]{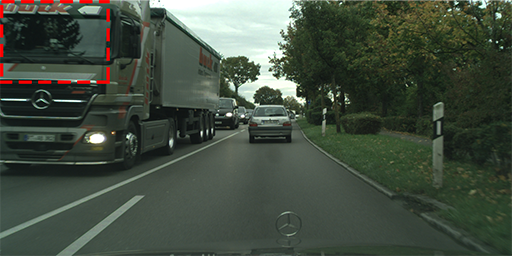}%
	}
	\hspace{0cm}
	\subfloat{\includegraphics[width=4.4cm]{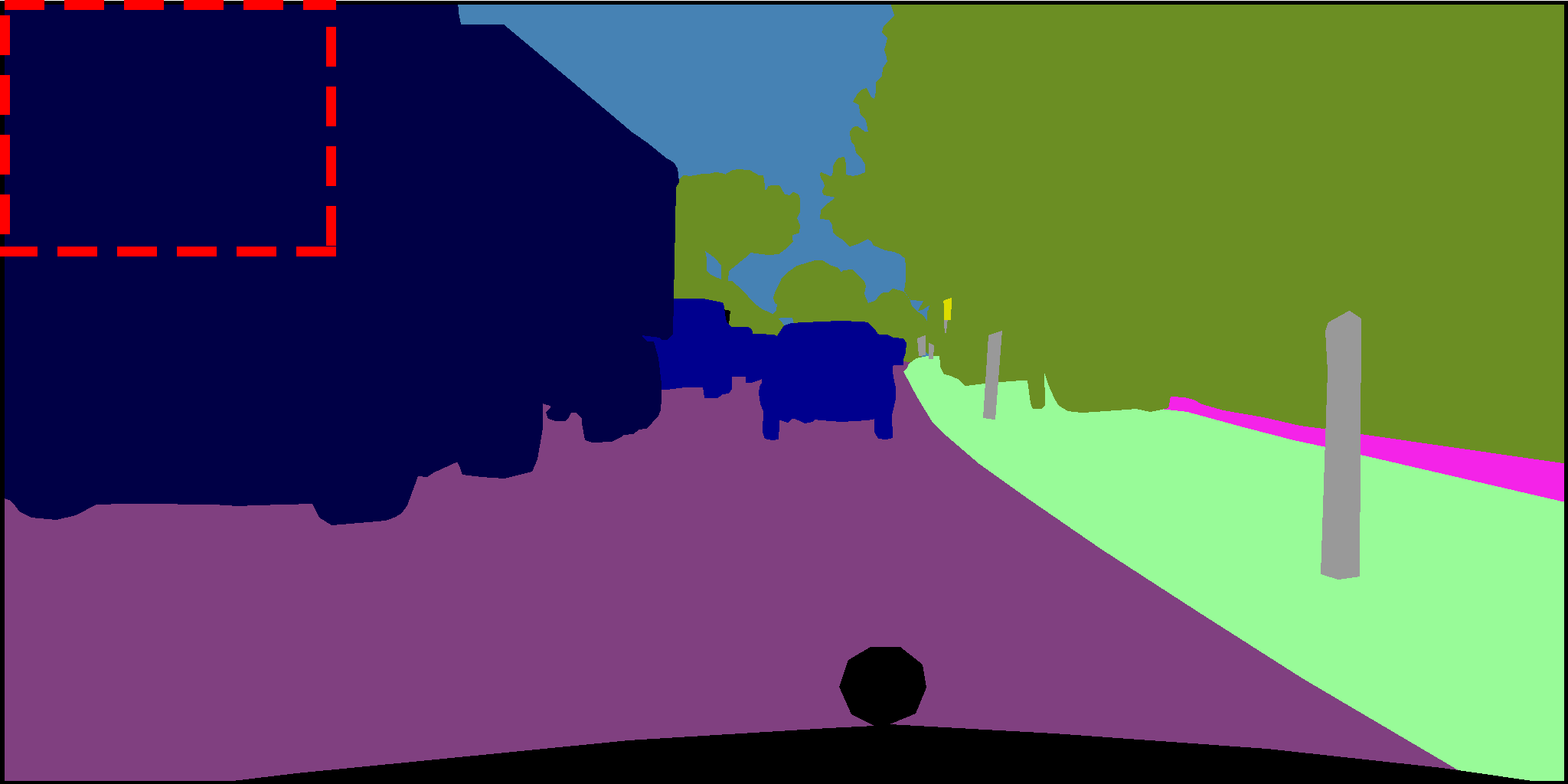}%
	}
	\hspace{0cm}
	\subfloat{\includegraphics[width=4.4cm]{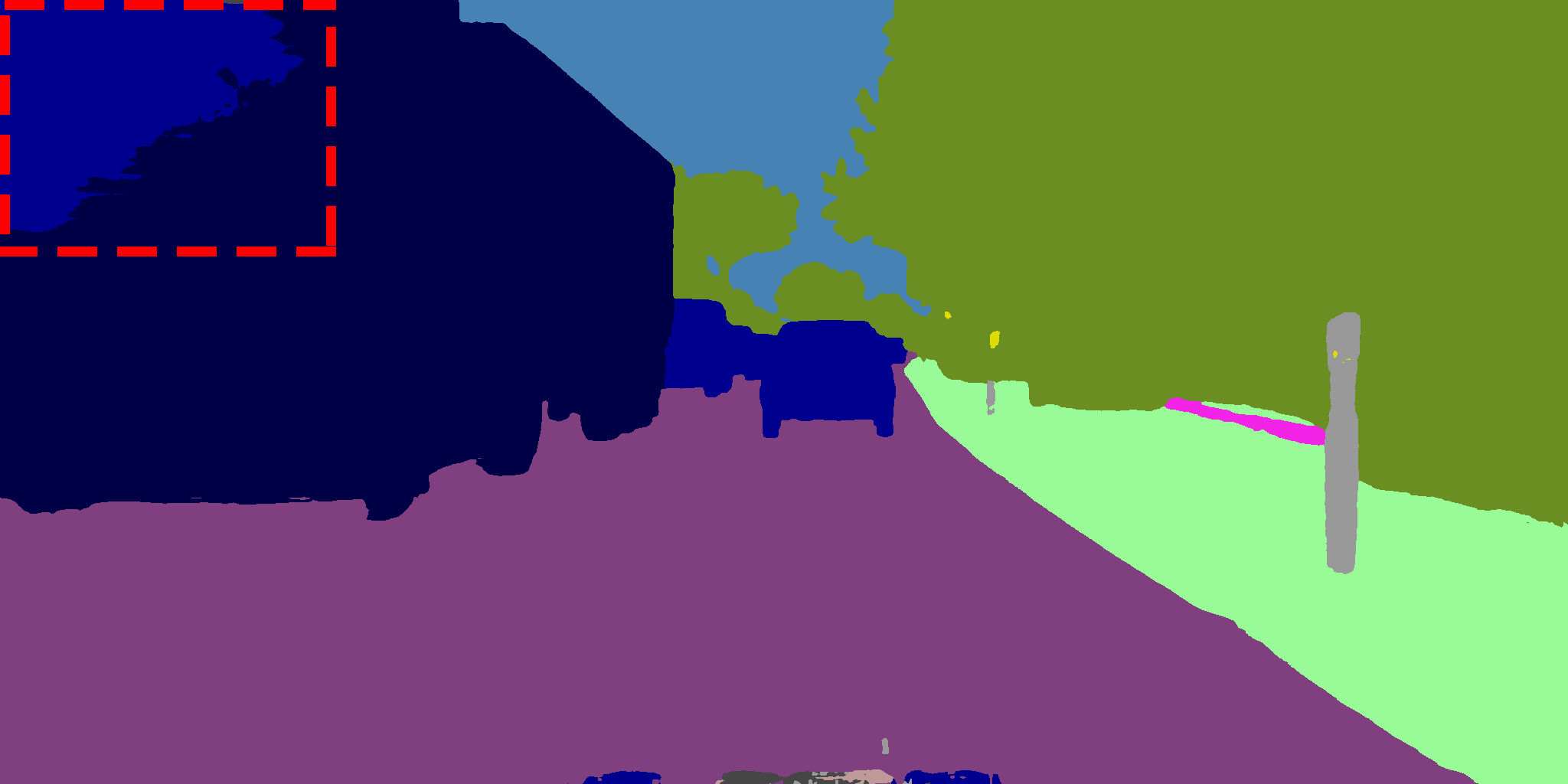}%
	}
	\hspace{0cm}
	\subfloat{\includegraphics[width=4.4cm]{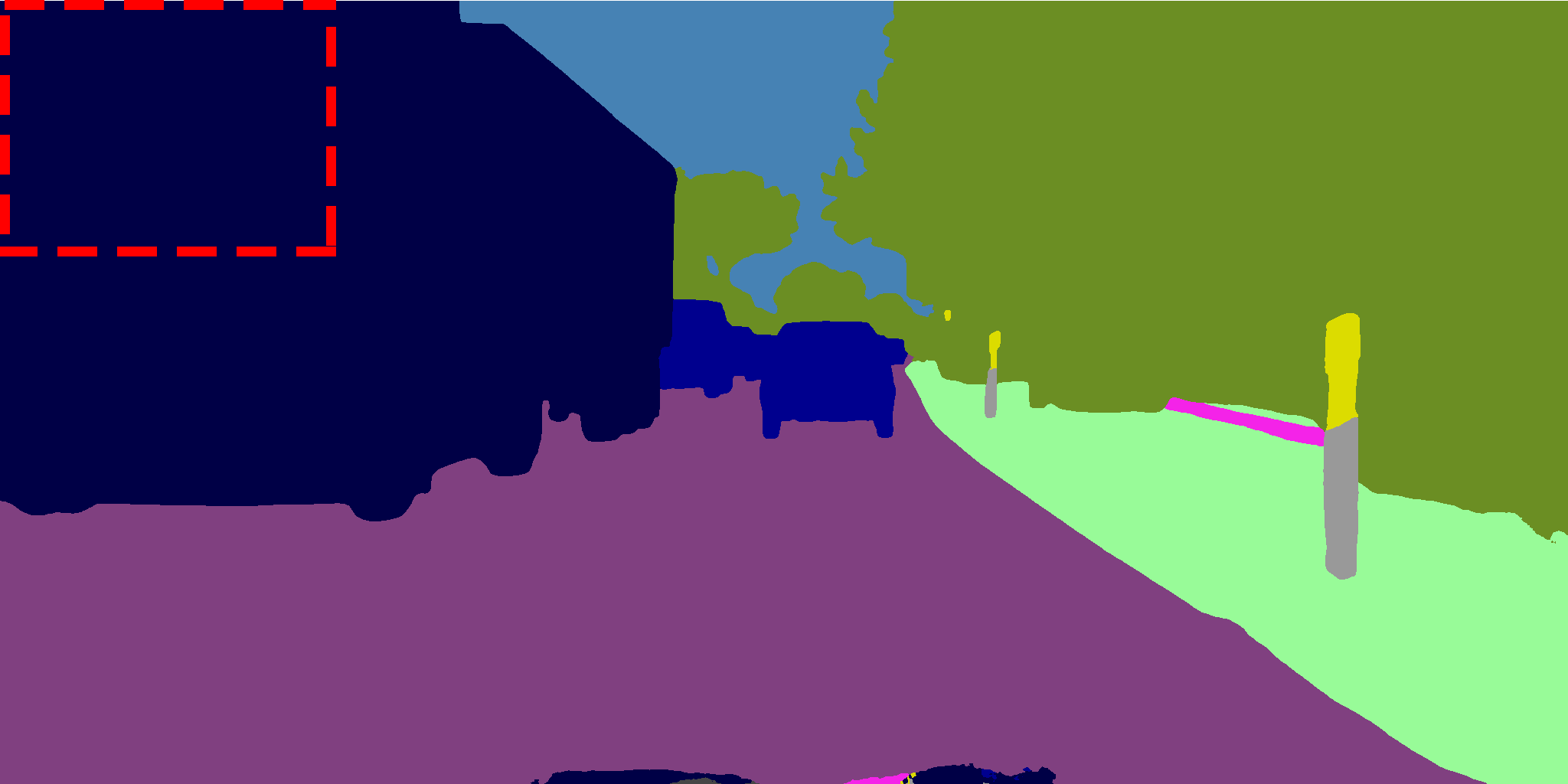}%
	}
	\vspace{0.15cm}
	\subfloat[{\scriptsize Image }]{\includegraphics[width=4.4cm]{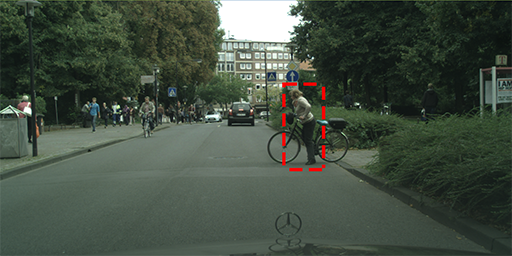}%
	}
	\hspace{0cm}
	\subfloat[{\scriptsize GT }]{\includegraphics[width=4.4cm]{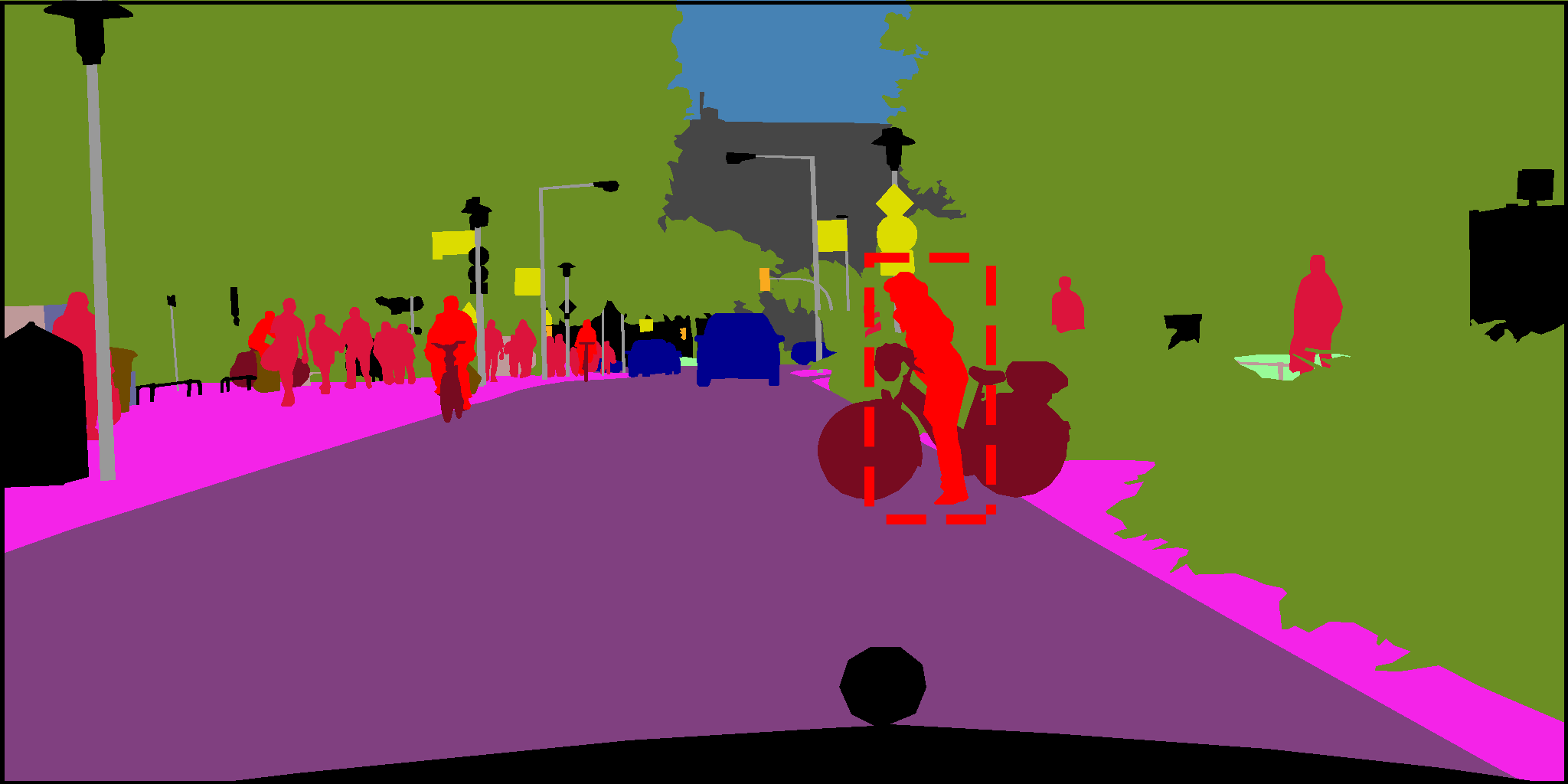}%
	}
	\hspace{0cm}
	\subfloat[{\scriptsize Baseline }]{\includegraphics[width=4.4cm]{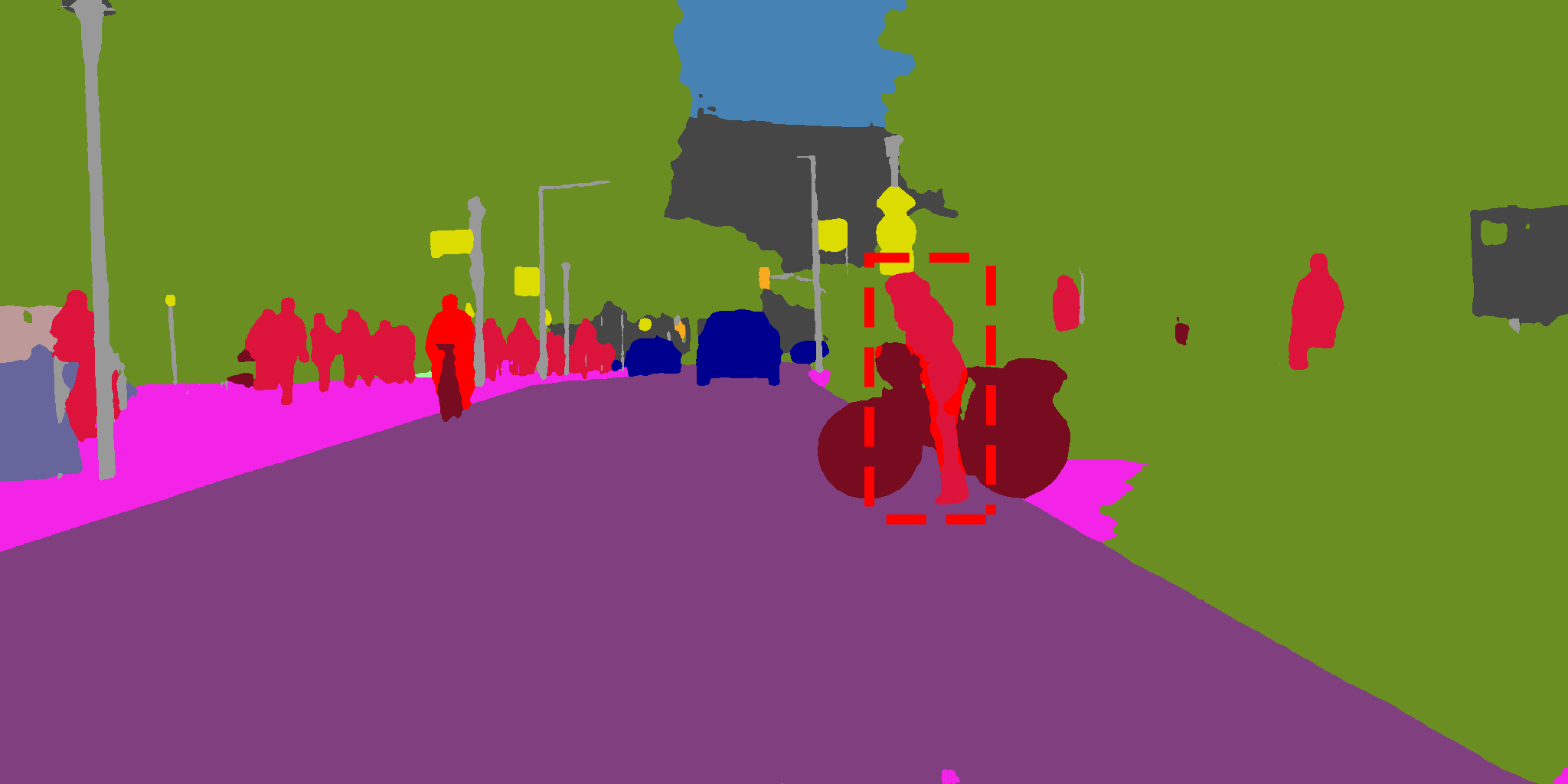}%
	}
	\hspace{0cm}
	\subfloat[{\scriptsize Ours }]{\includegraphics[width=4.4cm]{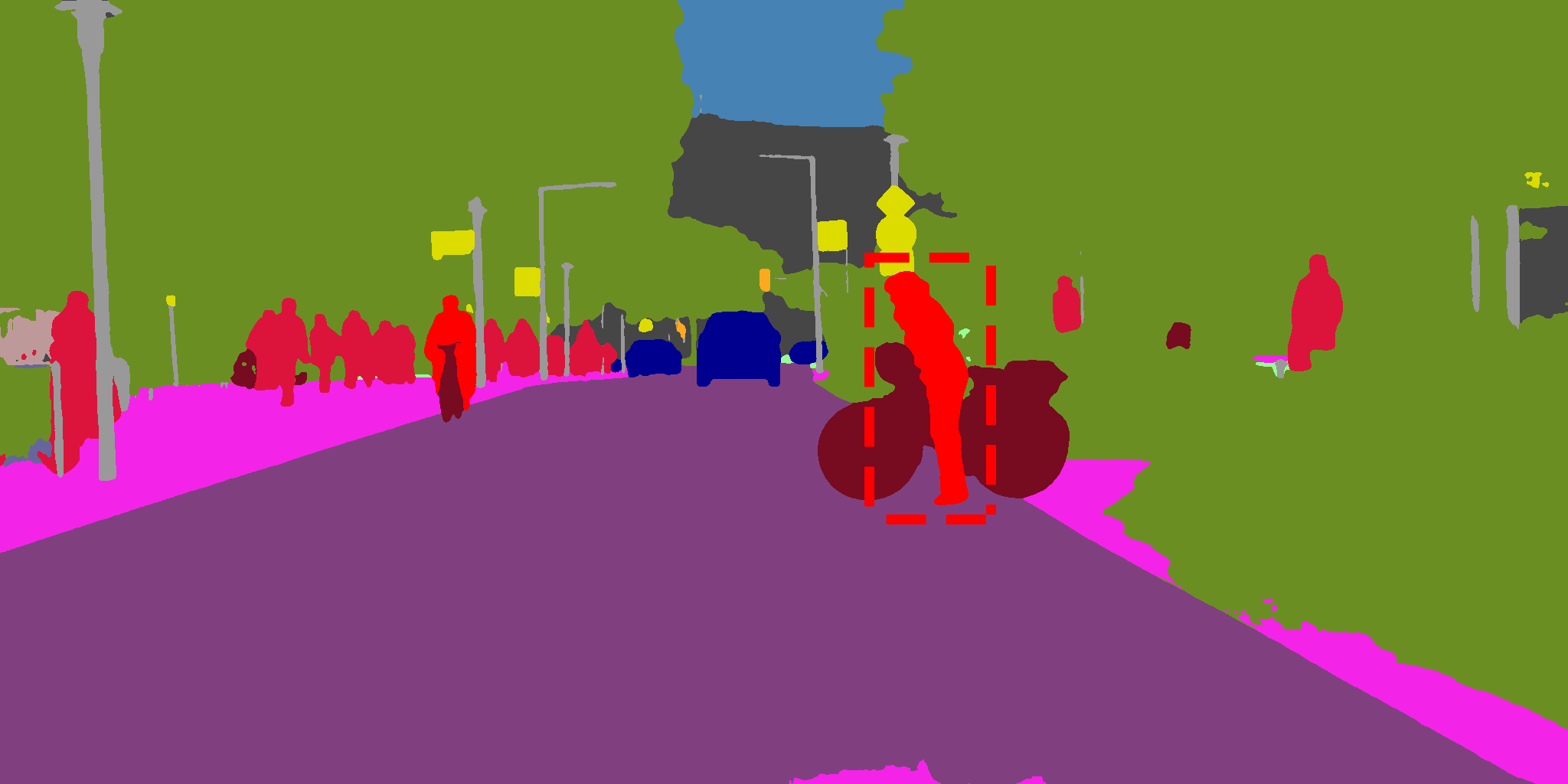}%
	}	
	\vspace{0.15cm}
	\subfloat {\includegraphics[width=18cm]{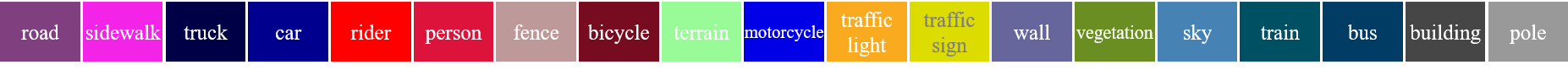}%
	}
	\caption{Qualitative results. Visual results of baseline and the proposed method on Cityscapes val set. The image from left to right is: input image, ground truth, prediction results from the baseline, prediction results from the proposed method. The improved areas are indicated by red dashed boxes.}
    \vspace{-0.2cm}
	\label{visualization_results}
\end{figure*}

\begin{figure}[tbp]
	\centering
	\captionsetup[subfloat]{labelsep=none,format=plain,labelformat=empty}
	
	\subfloat {\includegraphics[width=4.35cm]{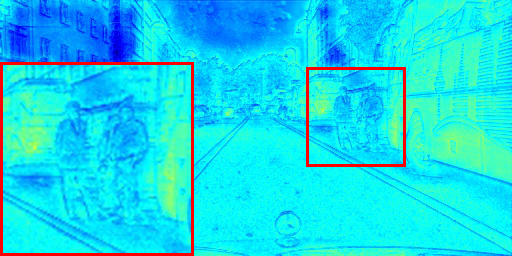}%
	}
	\hspace{0.01cm}
	\subfloat {\includegraphics[width=4.35cm]{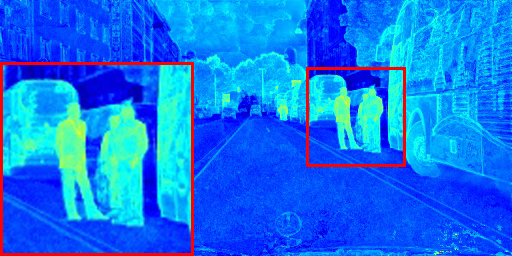}%
	}
	\vspace{0.15cm}
	\subfloat {\includegraphics[width=4.35cm]{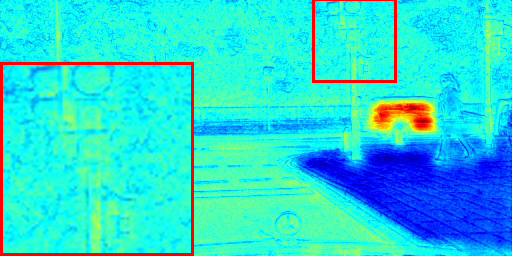}%
	}
	\hspace{0.01cm}
	\subfloat {\includegraphics[width=4.35cm]{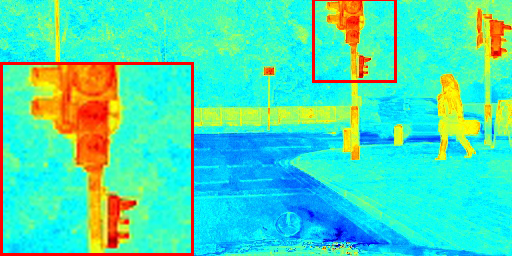}%
	}
	\vspace{0.15cm}
	\subfloat[{\scriptsize Bilinear }] {\includegraphics[width=4.35cm]{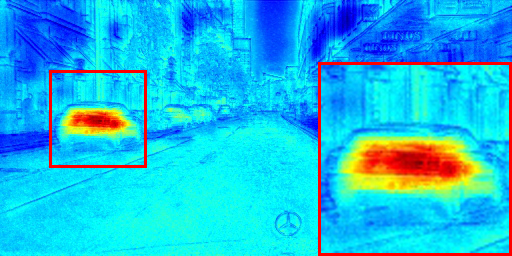}%
	}
	\hspace{0.01cm}
	\subfloat[{\scriptsize SRM}]  {\includegraphics[width=4.35cm]{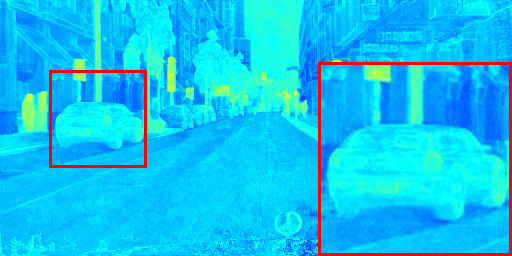}%
	}
	\caption{Visualization examples of the feature map $G_1$. \textbf{Left}: the bilinear upsampling method (Bilinear); \textbf{Right}: the designed Semantic Refinement Module (SRM).}
	\label{feature_map}
   \vspace{-0.5cm}
\end{figure}

\subsubsection{\bf Visualization} 
The qualitative results of the Cityscapes val set are displayed in Fig. \ref{visualization_results}. Compared with the baseline, the proposed method performs better segmentation results for some categories that are prone to be misclassified, such as the ``wall" and the ``fence" (in the second row), the ``car" and the ``truck" (in the third row). In the fourth rows, the ``rider" is misclassified as the ``person" by the baseline network due to the similar appearance. But the proposed method corrects the misclassified semantic labels successfully. In Fig. \ref{visualization_results}, the improved regions are indicated by red dashed boxes.
 
We also visualize the feature maps generated by the bilinear upsampling method and the designed SRM to illustrate the effectiveness of the learned offsets. The visualization examples on Cityscapes set are shown in Fig. \ref{feature_map}. The left column shows the feature maps obtained using the bilinear upsampling method, while the results generated by SRM are displayed in the right column. We enlarge the red box areas for a better observation. From Fig. \ref{feature_map}, it is observed that SRM recovers more clear boundaries on person, traffic sign, car and other kinds of objects, indicating the effectiveness of the learned offsets. 

\subsection{Experiment on Bdd100K}
TABLE \ref{bdd100k} presents the evaluated results on Bdd100K, further demonstrating the effectiveness of the proposed method. The mIoU is evaluated with single-scale inference. We set the image size to be $1280 \times 720$ for calculating the GFLOPs. The proposed method is compared with SOTA methods, including HANet \cite{HANet}, SFNet \cite{SFNet}, PFnet \cite{PFnet}, FRMSeg \cite{wang2023feature} and TScGAN \cite{Lateef2024ATC}. As shown in TABLE \ref{bdd100k}, the proposed method achieves 66.1\% mIoU on the Bdd100K val set with 59.5 GFLOPs. Even with a light-weight backbone, the proposed method achieves a better mIoU than TScGAN \cite{Lateef2024ATC} which employs the large Resnet-101 model as its backbone. Compared with our previous preliminary work \cite{wang2023feature} with the same backbone VAN-S, we further improve the 1.2\% mIoU with only 63\% of computational cost. 

\begin{table}[tbp]
	\centering
	\renewcommand\arraystretch{1.3}
	\caption{Experiments results on Bdd100K val set. The mIoU here is evaluated with single-scale inference. To calculate GFLOPs, the image resolution is set to $1280\times720$. ``G” and  ``P” stand for GFLOPs and Params respectively.}
	
	\begin{tabular}{@{}ccccc@{}}
		\toprule
		Method & Backbone   & mIoU(\%)$\uparrow$ &  G $\downarrow$  & P (M) $\downarrow$ \\ \midrule
		HANet \cite{HANet}  & ResNet-101 & 64.6    & -        & 65.4   \\
		SFNet \cite{SFNet} & ResNet-18  & 60.6     & 107.34   & 12.87  \\
		PFnet \cite{PFnet}  & ResNet-50   & 62.7     & 302.1    & 33.0  \\ 
		FRMSeg \cite{wang2023feature}  & VAN-S   & 64.9      & 94.6    & 16.5  \\ 
        \multirow{2}{*}{ TScGAN \cite{Lateef2024ATC}}   & ResNet-50  & 63.4    & -    & -  \\
                                                       & ResNet-101 X71 &  66.0   & -    & -  \\ \midrule
		\multirow{2}{*}{Ours}  & VAN-S        & {\bf 66.1}     & 59.5    & 15.4  \\
		                         & MSCAN-S        &  65.9     & 60.7    & 15.5  \\   \bottomrule
	\end{tabular}
	\label{bdd100k}
    \vspace{-0.2cm}
\end{table}

\begin{table}[tbp]
	\centering
	\renewcommand\arraystretch{1.3}
	\caption{Experiments results on ADE20K val set. The mIoU here is evaluated with single-scale inference. To calculate GFLOPs, the image resolution is set to $512\times512$.  ``G” and  ``P” stand for GFLOPs and Params respectively.}
	\begin{tabular}{ccccc}
		\toprule
		Method       & Backbone      & mIoU(\%)$\uparrow$  & G $\downarrow$  & P (M) $\downarrow$    \\ \midrule
		SFNet \cite{SFNet}       & ResNet-50     & 42.8     & 83.3   & 31.3  \\
		SegFormer \cite{SegFormer}   & MiT-B1        & 42.2     & 30.5   & 13.7  \\
		PVT  \cite{PVT}        & PVT-Small     & 39.8     & 44.5   & 28.2  \\
		SeMask-T \cite{SeMask} & SeMask Swin-T & 42.0     & 40     & 35    \\
		NRD  \cite{NRD}         & ResNet-101     & 44.0     & 49.0   &  - \\
		RTFormer  \cite{RTFormer}    & RTFormer-Base & 42.1     & 16.9   & 16.8   \\
		PRSeg-M \cite{PRSeg}     & ResNet-50     & 41.6     & 19.2   & 30     \\ 
        SegNeXt \cite{SegNeXt}     & MSCAN-S     & 44.3     & 15.9   & 13.9     \\

        SenFormer \cite{SenFormer}  & ResNet-50  & 44.4   & 179   &  55 \\ 
      ICPC \cite{Yu2023ICPCIP}   &ResNet-50    & 45.2     & 71.1   & - \\
        Vim \cite{Zhu2024VisionME}    & Vim-S   & 44.9   & -    & 46 \\ \midrule
        \multirow{2}{*}{Ours  }  & VAN-S        & 45.1     & 17.2    & 15.4 \\
		     & MSCAN-S        &  {\bf 45.2}    & 17.5    & 15.5  \\
 
		\bottomrule
	\end{tabular}
	\label{ade20k}
    \vspace{-0.2cm}
\end{table}
\subsection{Experiment on ADE20K}
The performance of the proposed method is also evaluated on the ADE20K val set. The compared methods include SFNet \cite{SFNet}, SegFormer \cite{SegFormer}, PVT \cite{PVT}, SeMask-T \cite{SeMask}, NRD \cite{NRD}, RTFormer \cite{RTFormer}, PRSeg-M\cite{PRSeg}, SegNeXt \cite{SegNeXt}, SenFormer \cite{SenFormer}, ICPC \cite{Yu2023ICPCIP}, Vim \cite{Zhu2024VisionME}. In the experiments, we adopt the same implementation settings on ADE20K as on other datasets. When calculating GFLOPs, the image size is configured as $512\times512$. The experimental results are shown in TABLE \ref{ade20k}. As can be observed from TABLE \ref{ade20k}, compared with other methods, the proposed method has the best performance. The proposed method achieves 45.2\% mIoU on the ADE20K val set with only 17.5 GFLOPs. Compared to SegNeXt \cite{SegNeXt}, the proposed method with the same MSCAN-S backbone achieves a 0.9\% increase in mIoU with similar computation cost and parameters. When comparing to the ICPC \cite{Yu2023ICPCIP} with a heavy ResNet-50, the proposed method with the light-weight MSCAN-S achieves same 45.2\% mIoU where only 25\% of computational cost is required.

\section{CONCLUSION}
In this paper, we propose the semantic and contextual refinement modules for semantic segmentation. At first, the semantic refinement module is designed to learn a transformation offset with the guidance of the high-resolution feature maps and the offsets in the neighbors' area, which significantly alleviate the misalignment problem brought by the upsampling of the low-resolution feature map. In addition, this paper builds the contextual refinement module to exploit the dependencies between pixels along with both spatial and channel dimensions, capturing global context information for semantic segmentation in an adaptive manner. To demonstrate the generality and effectiveness of our modules, this paper introduces the semantic refinement module and the contextual refinement module into different segmentation networks. The experimental results indicate that the proposed method achieves SOTA performance on three challenging datasets for semantic segmentation.

\section*{Acknowledgments}
The authors would like to thank Bingnan Han, Chaolin Lun and Xuanwu Yin from Xiaomi Technology for discussing the light-weight network design.

This work is partially supported by the China Natural Science Foundation (NSFC, No. 62301497, No. 62101503 and No. 62372150), the project of the Ministry of Science and Technology (No. G2023019011L), the Henan Key Research and Development Program (No. 231111212000), and the Science and Technology Project of Henan Province under Grant (No. 242102211017).

\bibliographystyle{IEEEtran}
\bibliography{IEEEabrv,strings} 
\end{document}